%% file: neurips_2026.tex
\documentclass{article}

\PassOptionsToPackage{numbers,compress}{natbib}

\usepackage[main,final]{neurips_2026}

\usepackage[utf8]{inputenc} 
\usepackage[T1]{fontenc}    
\usepackage{hyperref}       
\usepackage{url}            
\usepackage{booktabs}       
\usepackage{amsfonts}       
\usepackage{nicefrac}       
\usepackage{microtype}      
\usepackage{xcolor}         

\usepackage{graphicx}
\usepackage{wrapfig}
\usepackage{algorithm}
\usepackage{algorithmic}
\usepackage{amsmath}
\usepackage[table]{xcolor}

\usepackage{booktabs}
\usepackage{graphicx}

\usepackage{multirow}       
\DeclareMathOperator*{\argmax}{arg\,max}
\usepackage{enumitem}

\title{GOLD: Geometric Optimized Latent Diffusion for Structure-Aware RNA Inverse Folding}

\author{%
  David S.~Hippocampus\thanks{Use footnote for providing further information
    about author (webpage, alternative address)---\emph{not} for acknowledging
    funding agencies.} \\
  Department of Computer Science\\
  Cranberry-Lemon University\\
  Pittsburgh, PA 15213 \\
  \texttt{hippo@cs.cranberry-lemon.edu} \\
}

\author{%
  \textbf{Qi Si}$^{1}$%
  \thanks{
    contributed equally to this work.}
  \quad
  \textbf{Xuyang Liu}$^{1}$\footnotemark[1]
  \quad
  \textbf{Penglei Wang}$^{2}$\footnotemark[1]
  \quad
  \textbf{Shuo Su}$^{3}$
  \quad
  \textbf{Limei Han}$^{4}$
  \\[2pt]
  \textbf{Mengzhu Li}$^{4}$
  \quad
  \textbf{Xin Guo}$^{1,4}$%
  \thanks{Corresponding to
   \texttt{guoxin@sais.org.cn}
    and \texttt{cheng\_yuan@fudan.edu.cn}.}
  \quad
  \textbf{Yuan Qi}$^{1,4,5}$
  \quad
  \textbf{Yuan Cheng}$^{1,4}$\footnotemark[2]
  \\[5pt]
  {\small
  $^{1}$Shanghai Academy of Artificial Intelligence
  for Science}
  \\
  {\small
  $^{2}$School of Biomedical Engineering,
  Shanghai Jiao Tong University}
  \\
  {\small
  $^{3}$Community Science Department, Kuaishou}
  \\
  {\small
  $^{4}$Artificial Intelligence Innovation and
  Incubation Institute, Fudan University}
  \\
  {\small
  $^{5}$Zhongshan Hospital, Fudan University}
}

\begin{document}

\maketitle

\begin{abstract}
  Designing RNA sequences that fold into specific 3D structures is a fundamental challenge in therapeutics. Current RNA inverse folding methods typically condition on 3D backbone coordinates but the generation is limited to 1D discrete sequences. This ignores the continuous side-chain orientations, which are essential for overall structural stability. To address this issue, we explicitly introduce side-chain geometry into the RNA generative process. Unlike co-design frameworks that diffuse across separate raw spaces, we propose Geometric Optimized Latent Diffusion (\textbf{GOLD}), which project sequence and structure into a unified latent manifold. GOLD employs a novel GeoVAE to elegantly integrate discrete RNA sequences with continuous side-chain angles. By mapping these modalities into a single, smooth latent manifold, GOLD naturally facilitates a synergistic co-diffusion process. To enhance structural validity, we further introduce a training-free, gradient-guided sampling mechanism. By decoding the generated sequence probabilities and side-chain geometries into structures during the sampling process, we compute analytical gradients from physical constraints. This dynamic feedback actively guides the generated sequence to a physically stable state. Extensive experiments validate the superiority of GOLD, achieving state-of-the-art performance across sequence recovery, secondary structure validity, and 3D foldability metrics. Our code and data are publicly available at
\url{https://github.com/darkflash03/GOLD}.
\end{abstract}

\section{Introduction}
\label{introduction}

RNA is essential in modern therapeutics and synthetic biology, driving key innovations such as mRNA vaccines and RNA aptamers ~\citep{damase2021rna, wong2024deep}. Since the biological function of RNA molecules depends largely on their 3D structures, the development of new RNAs requires the design of sequences that can reliably fold into specific target structures. This task of designing a nucleotide sequence for a target 3D structure is known as RNA inverse folding ~\citep{das2010automated, GarcaMartn2013RnaifoldAC}. Driven by deep learning, current generative models typically condition on the 3D coordinates of the RNA backbone to predict discrete tokens in 1D sequence space ~\citep{tan2024rdesign, joshi2025grnade}. However, the spatial orientations of nucleobase side-chains directly dictate the formation of base-pairing and stacking interactions in RNA, which ultimately govern the stability of the entire 3D structure \citep{leontis2001geometric, murray2003rna}. Due to this geometric dependency, there are inherent limitations in restricting the generation of targets to 1D sequence space. If the orientation of the side chain in 3D space is not explicitly modeled, the generated sequences may fail to satisfy spatial constraints, resulting in atomic collisions within the generated structure.

Beyond 1D sequence generation, a broader paradigm in macromolecular modeling attempts to jointly capture sequences and 3D geometries. Pioneered in protein engineering ~\citep{Lisanza2023Joint, chen2025allatom} and recently extended to \textbf{de novo} RNA design ~\citep{ma2025riboflow, li2026a}, these frameworks highlight the importance of modeling both modalities. However, a fundamental distinction exists in the task objective: while \textbf{de novo} co-design generates both sequence and structure from scratch, inverse folding operates under the physical constraints of a predefined 3D backbone. To optimally leverage side-chain geometry within this rigidly constrained setting, we propose Geometric Optimized Latent Diffusion (GOLD). GOLD employs a Geometry-aware Variational Autoencoder (GeoVAE) to elegantly map discrete RNA sequences and continuous side-chain angles into a unified, smooth latent manifold. We further refine this manifold by aligning it with embeddings from a pre-trained RNA foundation model, leveraging its sequence representations to enrich our latent space. By executing the diffusion process directly within this shared and enriched space, our model inherently couples sequence identity with 3D geometry, ensuring a natural and consistent alignment of physical constraints during sequence generation.

In addition to a unified representation, maintaining the physical consistency of the generated structures is essential. To achieve this goal, current state-of-the-art models \citep{si2026structure, hu2026rider} often utilize Reinforcement Learning (RL) \citep{black2024training, Fan2023DPOK, wang2025finetuning} to fine-tune their output based on external folding tools, which typically requires significant computational resources. To provide a more efficient alternative, GOLD leverages explicit sequence and continuous side-chain angles to introduce a training-free, gradient-guided sampling mechanism. During the denoising process, our method calculates gradients directly from 2D pairing rules and 3D collision constraints to guide the sequence toward physically stable states without the need for additional post-training.

Our main contributions are summarized as follows:
\begin{itemize}
\item \textbf{Explicit Geometric Awareness in RNA Inverse Folding:} We identify continuous side-chain orientation as the critical missing link in current inverse folding methods, and introduce continuous side-chain orientations as a generative target in our method.

\item \textbf{Unified Representation for RNA Co-generation:} We introduce a novel generative framework that combines GeoVAE and Latent Diffusion Model (LDM). GeoVAE compresses discrete sequences and continuous side-chain angles into a unified latent manifold, enabling the LDM to perform physically consistent sequence generation.

\item \textbf{Training-Free Gradient Guidance:} Based on our continuous geometric representation, we introduce a sampling mechanism that incorporates analytical gradients from 2D pairing and 3D collision constraints.

\item \textbf{Comprehensive Benchmark Performance:} GOLD achieves state-of-the-art results across a comprehensive suite of 1D, 2D, and 3D metrics. Our model simultaneously improves sequence recovery and physical foldability, demonstrating that explicit geometric modeling leads to more reliable RNA designs.
\end{itemize}

\section{Related Work}
\label{related_works}

\subsection{RNA Inverse Folding}
Historically, computational modeling of RNA inverse folding began as a thermodynamic optimization problem. Early physics-based methods ~\citep{hofacker1994fast, zadeh2011nupack, busch2006info, andronescu2004new} framed sequence design as a task to minimize free-energy through dynamic programming and heuristic search. As the field advanced, deep learning architectures began to replace traditional algorithms. Initial data-driven strategies primarily relied on reinforcement learning and autoregressive models ~\citep{runge2018learning}. To better capture structural dependencies, the field subsequently shifted to Graph Neural Networks (GNNs) ~\citep{tan2024rdesign, joshi2025grnade, tan2025r3design}, which actively leverage spatial message passing in 3D coordinates. Concurrently, RNA foundation models ~\citep{chen2022interpretable, penic2024_rinalmo, Brixi2026} were developed to capture deep co-evolutionary semantics. Most recently, following breakthroughs in protein design ~\citep{Chroma2023, yim2023se3}, diffusion frameworks were adapted for RNA design, including discrete generative diffusion ~\citep{han2024ribodiffusion} and hyperbolic discrete diffusion ~\citep{hou2025hyperbolic}. To enhance structural fidelity, reinforcement learning ~\citep{si2026structure, hu2026rider} is currently integrated to fine-tune these diffusion models based on external folding rewards. However, these refinement approaches typically require significant computational requirements.

\subsection{Latent Diffusion and Controllable Generation}

Latent Diffusion Models (LDMs) manage the complexity of discrete or mixed data spaces by compressing information into continuous, low-dimensional manifolds, based on foundational autoencoders \citep{kingma2013auto, higgins2017betavae, oord2017neural}. Operating within a continuous diffusion framework \citep{ho2020denoising, song2021scorebased} provides a smooth and differentiable optimization landscape while reducing computational overhead. This formulation enables controllable generation through gradient-based guidance \citep{chung2023diffusion, qiu2023freedom, zhao2022egsde}, which guide the reverse sampling trajectory by injecting gradients from auxiliary energy functions in each denoising step. Based on these advances, GOLD maps discrete RNA sequences and side-chains into a shared continuous latent space. This representation supports a training-free sampling process, using gradients from 2D pairing and 3D collision constraints to guide the model toward physically valid structures.

\section{Preliminaries}
\label{preliminary}

\subsection{Geometric Representation of RNA}
\begin{wrapfigure}{r}{0.5\textwidth}
  \vspace{-15pt}
  \centering
  \includegraphics[width=\linewidth]{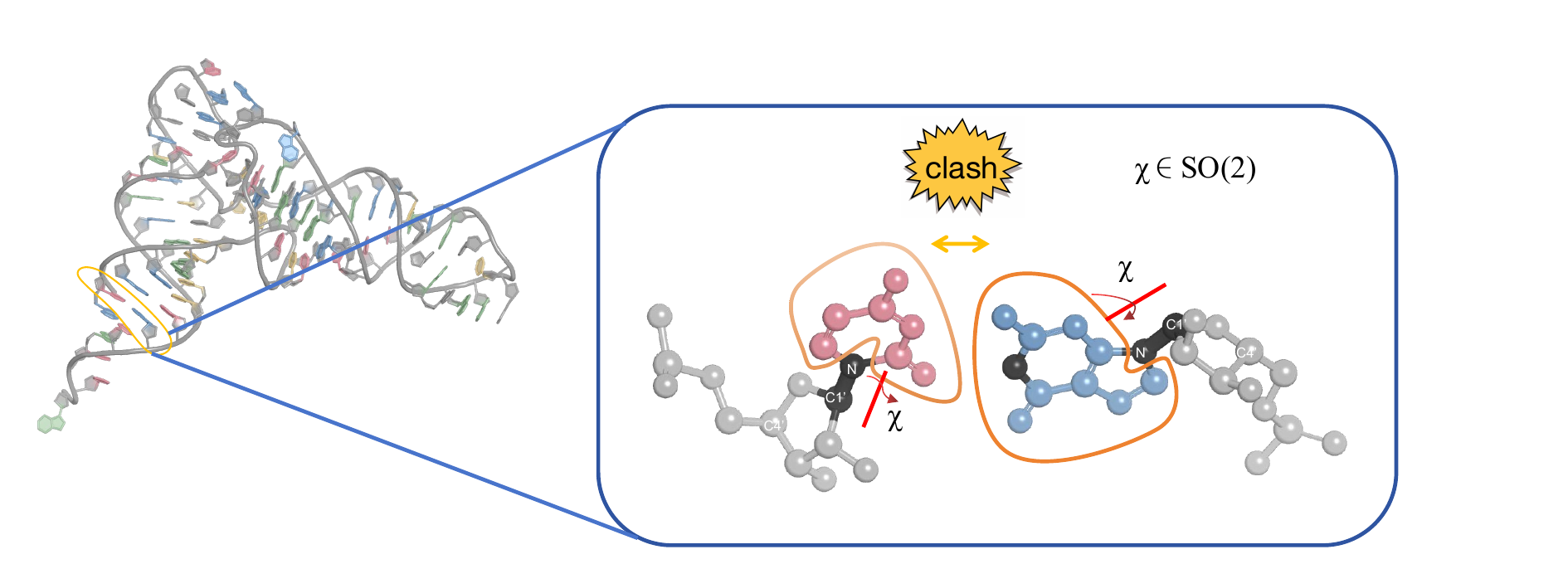 } 
  \caption{\textbf{RNA geometry.} The backbone establishes local frames ($\mathbf{R}_{bb}$), while the side-chain orientation is governed by the glycosidic torsion angle ($\chi \in \mathrm{SO}(2)$). Modeling this angle in 3D space helps identify potential steric clashes between rigid base planes.}
  \label{fig:motivation}
  \vspace{-10pt}
\end{wrapfigure}

Without explicit awareness of physical volume, RNA inverse folding models frequently produce discrete sequences that suffer from steric clashes when translated into 3D structures (Figure~\ref{fig:motivation}). Structurally, the RNA nucleobases exhibit a rigid, planar geometry inherent to their aromatic rings ~\citep{abramson2024accurate}. Since sequence identity typically defines this internal shape, the nucleobase's motion can be effectively approximated as a single rotation around the glycosidic bond. Consequently, this limitation establishes the glycosidic torsion angle $\chi$ as the primary structural degree of freedom (DoF) defining the spatial orientation and collision boundaries of the side-chains ~\cite{baek2024accurate}.

To formalize this constraint, we note that while the global position and orientation of a side-chain are represented in $\mathrm{SE}(3)$, its DoF under the rigid-body assumption is restricted to a 1D rotation around the glycosidic bond axis, represented by $\mathrm{SO}(2)$. Here, we define the Kinematic Approximation Error (\textbf{KAE}) as the root-mean-square deviation (RMSD) between the native atomic coordinates of the side-chain and those recovered jointly using the standard rigid base templates, the backbone coordinates and the side-chain angles $\chi$. To evaluate this physical prior, we analyzed approximately 8,000 native RNA structures. Our results demonstrate that retaining only the angle $\chi \in \mathrm{SO}(2)$ as a 1D continuous DoF is sufficient to bound the KAE below 0.2 \text{\AA}.

Based on this insight, we define a deterministic $\mathrm{SE}(3)$-equivariant kinematic building operator $\mathcal{B}$. Given the coarse-grained coordinates of the backbone $\mathbf{X}_{bb} \in \mathbb{R}^{N \times 3 \times 3}$, we extract local rotation matrices $\mathbf{R}_{bb}^{(i)} \in \mathrm{SO}(3)$ and translation vectors $\mathbf{T}_{bb}^{(i)} \in \mathbb{R}^3$ from anchor atoms. Assuming standard rigid-body templates $\mathbf{T}_{base}^{(i)} \in \mathbb{R}^{4 \times 14 \times 3}$, the side-chain coordinates $\mathbf{X}_{sc}^{(i)}$ can be reconstructed as:
\begin{equation}
\mathbf{X}_{sc}^{(i)} = \mathcal{B}(\mathbf{S}_i, \chi_i, \mathbf{X}_{bb}^{(i)}) = \left( \mathbf{S}_i \cdot \mathbf{T}_{base} \right) \mathbf{R}_{\chi_i} \mathbf{R}_{bb}^{(i)} + \mathbf{T}_{bb}^{(i)}
\label{eq:template}
\end{equation}

where $\mathbf{S}_i$ is the predicted sequence probability and $\mathbf{R}_{\chi_i} \in \mathrm{SO}(3)$ denotes the 3D rotation matrix constructed from the 1D angle $\chi_i \in \mathrm{SO}(2)$ around the local glycosidic bond axis. By formalizing this geometric relationship, we not only capture the intrinsic coupling between sequence probabilities and side-chain geometry but also make $\mathcal{B}$ as a differentiable operator. This lays the groundwork for active physical interventions: for instance, when spatial overlaps occur during generation, $\mathcal{B}$ translates physical repulsion into analytical gradients that backpropagate to refine both sequence probabilities and torsional orientations. We detail this mechanism in Appendix~\ref{app:motivation}.

\subsection{Problem Formulation}
Given a target RNA backbone $\mathbf{X}_{bb}$ of length $N$, our primary objective is to generate a physically valid discrete sequence $\mathbf{S} \in \mathcal{A}^N$, where $\mathcal{A} = \{\text{A, U, C, G}\}$ is the nucleotide alphabet. Traditional inverse folding models often treat this as a purely token prediction task \cite{ingraham2019generative}. This limits their ability to perceive geometric information such as side-chain collisions. As the analysis above suggests, the physical validity of a sequence is significantly influenced by the side chain torsion angle $\chi$. Therefore, the core problem we seek to address is: \textbf{How can side-chain angles serve as geometric constraints to guide physically valid sequence generation?}

\section{Methods}
\label{methods}

\begin{figure*}[t]
    \centering
    \includegraphics[width=\textwidth]{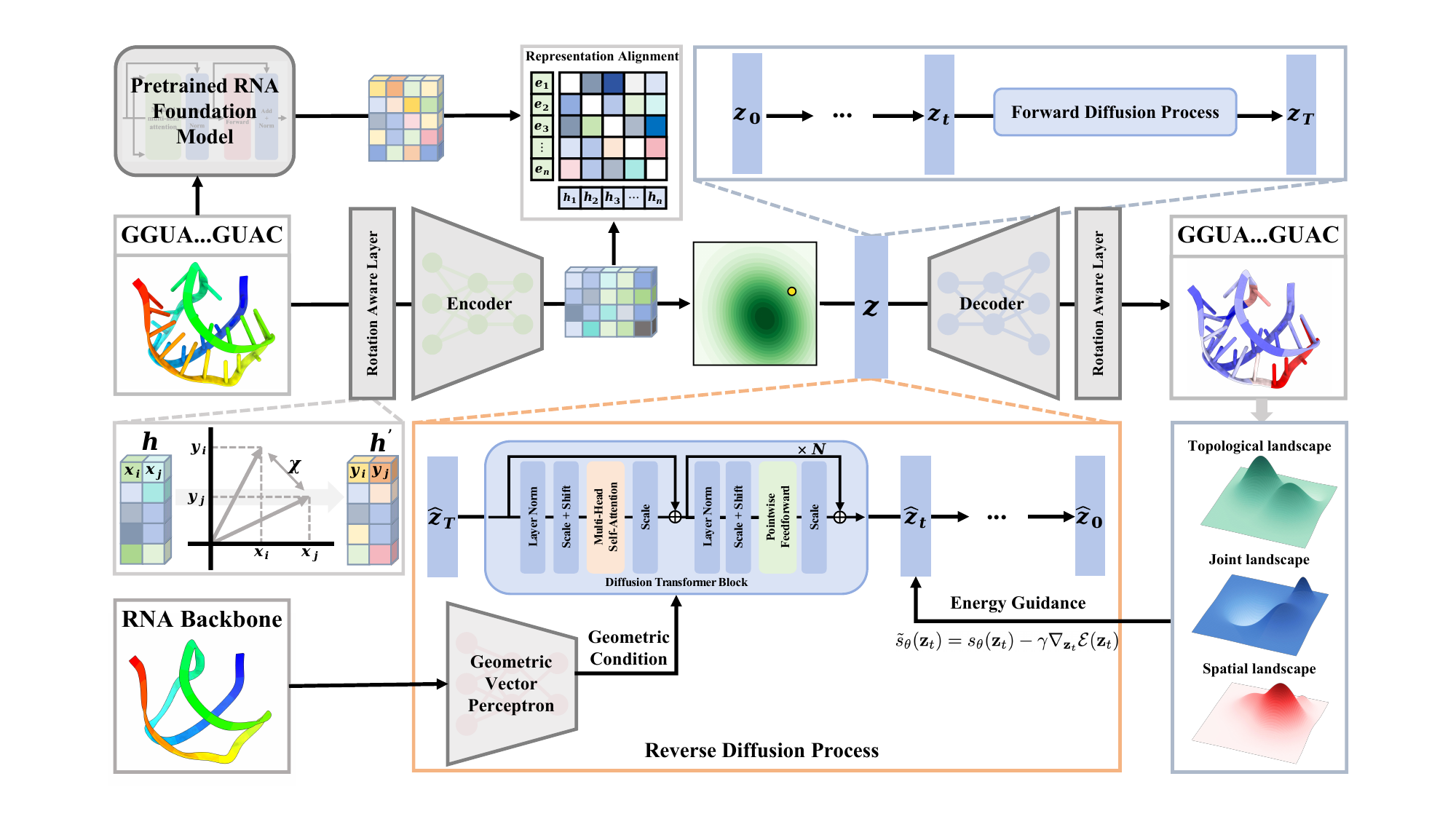}
    \caption{
    GOLD architecture for RNA inverse folding task. GOLD first encodes RNA sequences and structures into a unified latent space using an rotation aware (SO(2)) encoder, then decodes the latent embeddings into the raw space. Subsequently, a score-based diffusion model, enhanced by a geometric vector perceptron module, is employed in the latent space for RNA generation. During inference, GOLD utilizes an energy-based guidance mechanism to generate desired RNA samples.
    }
    \label{fig:framework}
\end{figure*}

\subsection{Overall Framework}
\label{framework}
As illustrated in Fig ~\ref{fig:framework}, we propose GOLD, a geometry-aware latent diffusion framework for RNA inverse folding. GOLD consists of three stages: (1) a Geometric Variation Autoencoder (\textbf{GeoVAE}) maps discrete sequences and torsion angles into a continuous latent space; (2) a latent diffusion model captures the distribution of valid RNA designs conditioned on 3D backbone geometry; and (3) a differentiable energy-guided sampling procedure guides reverse denoising to generate spatially feasible and topologically consistent samples.

\subsection{Geometric Variational Autoencoder (GeoVAE)}
\label{GeoVAE}
To process heterogeneous sequence tokens and geometric angles while enabling interaction with the 3D physical space, we design an $\mathrm{SO}(2)$-equivariant GeoVAE.

\textbf{Rotation-Aware Encoding and Decoding.} 
Let the dense features of the discrete sequence $\mathbf{S} \in \{0, 1\}^{N \times 4}$ be $\mathbf{H} \in \mathbb{R}^{N \times D}$. To eliminate $2\pi$ periodicity, continuous torsion angles $\boldsymbol{\chi}$ are encoded as unit circle vectors $c_i = [\cos \chi_i, \sin \chi_i]^\top$. As shown in the bottom left of Figure~\ref{fig:framework}, during encoding process, the \textbf{Rotation Aware Layer} reshapes sequence features $\mathbf{h}_i$ into complex pairs and applies $\mathrm{SO}(2)$ rotations:
\begin{equation}
\begin{bmatrix} x'_{i,k} \\ y'_{i,k} \end{bmatrix} = \mathbf{R}(\chi_i) \begin{bmatrix} x_{i,k} \\ y_{i,k} \end{bmatrix} = \begin{bmatrix} \cos \chi_i & -\sin \chi_i \\ \sin \chi_i & \cos \chi_i \end{bmatrix} \begin{bmatrix} x_{i,k} \\ y_{i,k} \end{bmatrix}
\end{equation}
Following this rotation-aware integration, the coupled features pass through multiple Transformer encoder layers, which ultimately map the inputs into the continuous latent distribution $q_{\phi}(\mathbf{z}|\mathbf{S}, \boldsymbol{\chi}) \sim \mathcal{N}(\boldsymbol{\mu}, \boldsymbol{\sigma}^2)$. During decoding, the sampled latent variable $\mathbf{z}$ is decoupled through the rotation-aware layer~\citep{worrall2017harmonic, thomas2018tensor}: 
\begin{itemize}
    \item \textbf{Sequence}: We decode the probability using the magnitude: $\hat{\mathbf{s}}_i = \text{MLP}_{seq}(\sqrt{x^2 + y^2})$.
    \item \textbf{Torsion}: To maintain sensitivity to spatial orientation, the angle is decoded directly from the complete phase information: $\hat{\chi}_i = \text{Normalize}(\text{MLP}_{angle}(\mathbf{z}_i))$.
\end{itemize}

\textbf{Joint Optimization.} We feed the predicted soft probabilities $\hat{\mathbf{S}}$ and angles $\hat{\boldsymbol{\chi}}$ directly into the operator $\mathcal{B}$ defined in Equation (~\ref{eq:template}). Combined with the fixed target backbone $\mathbf{X}_{bb}$, this will output the 3D coordinates of the side chain $\hat{\mathbf{X_{sc}}}$. This operation establishes a differentiable pathway from the latent space $\mathcal{Z}$ to the physical space. The GeoVAE is optimized end-to-end via the following joint objective:
\begin{equation}
\mathcal{L}_{GeoVAE} = \lambda_1 \mathcal{L}_{seq} + \lambda_2 \mathcal{L}_{angle} + \lambda_3 \mathcal{L}_{coord} + \lambda_4 \mathcal{L}_{repa} + \lambda_5 \mathcal{L}_{KL}
\end{equation}
where each term provides explicit supervision: 
\textbf{(1) $\mathcal{L}_{seq}$ (Cross-Entropy)} minimizes the discrepancy between the predicted soft probabilities $\hat{\mathbf{S}}$ and the ground-truth one-hot sequence $\mathbf{S}$; 
\textbf{(2) $\mathcal{L}_{angle}$ (MSE)} computes the reconstruction error between the predicted continuous 2D torsion vectors and the native vectors; 
\textbf{(3) $\mathcal{L}_{coord}$ (MSE)} directly penalizes the spatial deviation between the reconstructed side-chain coordinates $\hat{\mathbf{X}}_{sc}$ and the native side-chain coordinates $\mathbf{X}_{sc}$; 
\textbf{(4) $\mathcal{L}_{repa}$ (Cosine Similarity)} computes $1 - \cos(\mathbf{z}, \mathbf{z}_{FM})$ to align the latent variable $\mathbf{z}$ with the embeddings of pre-trained foundation model $\mathbf{z}_{FM}$~\cite{yu2025repa}; and 
\textbf{(5) $\mathcal{L}_{KL}$ (KL Divergence)} regularizes the approximate posterior $q_{\phi}(\mathbf{z}|\mathbf{S}, \boldsymbol{\chi})$ to conform to the standard normal prior $\mathcal{N}(\mathbf{0}, \mathbf{I})$.

\subsection{Latent Diffusion}
\label{LDM}
After establishing the continuous manifold $\mathcal{Z}$, the sequence generation task is formulated as a diffusion process within this continuous space, avoiding the complex transition matrix optimization typically required in discrete diffusion.

\textbf{Forward and Reverse Diffusion Processes.} Given a clean latent representation $\mathbf{z}_0 \sim \mathcal{Z}$, the forward diffusion process progressively injects Gaussian noise according to a predefined noise schedule $\bar{\alpha}_t$:
\begin{equation}
q(\mathbf{z}_t | \mathbf{z}_0) = \mathcal{N}(\mathbf{z}_t; \sqrt{\bar{\alpha}_t}\mathbf{z}_0, (1 - \bar{\alpha}_t)\mathbf{I})
\end{equation}
During the reverse denoising process, the generative distribution is conditioned on the target backbone $\mathbf{X}_{bb}$. We construct a spatial graph and utilize Geometric Vector Perceptrons (GVP)~\cite{jing2021learning} to extract rotation-invariant scalars $\mathrm{S}$ and $\mathrm{SE}(3)$-equivariant vectors $\mathbf{V}$:
\begin{equation}
\mathbf{S}' = \text{MLP}_s \Big( \big[ s \parallel \|\mathbf{W}_h \mathbf{V}\|_2 \big] \Big), \quad \mathbf{V}' = \mathbf{W}_v (\mathbf{W}_h \mathbf{V}) \odot \sigma \big( \text{MLP}_{gate}(s') \big)
\end{equation}
where $\mathbf{W}_h$ and $\mathbf{W}_v$ are learnable parameters. As illustrated in Figure ~\ref{fig:framework}, these updated representations ($\mathrm{S}$, $\mathrm{V}$) collectively constitute the \textbf{Geometric Condition}, providing the essential 3D structural scaffold to modulate the subsequent diffusion process.

\textbf{Model Optimization.} These aggregated geometric conditions, along with timestep embeddings, modulate the hidden states of the Diffusion Transformer (DiT)~\citep{Peebles2022DiT} via AdaLN mechanisms. With the GeoVAE parameters frozen, the DiT is trained to predict the clean latent $\mathbf{z}_0$ by minimizing the mean squared error:
\begin{equation}
\mathcal{L}_{Diff} = \mathbb{E}_{\mathbf{z}_0, \epsilon, t} \left[ \| \mathbf{z}_\theta(\mathbf{z}_t, \mathbf{X}_{bb}, t) - \mathbf{z}_0 \|_2^2 \right]
\end{equation}

\subsection{Dual-Guided Sampling}
\label{dual_guide_sampling}

Although the diffusion model effectively captures the global data distribution, generating physically feasible RNAs requires explicitly resolving conflicts to ensure collision-free and topologically correct structures. Leveraging the differentiable pathway established by operator $\mathcal{B}$, we introduce a dual-guided mechanism to actively direct the sampling process.

\textbf{Energy Guidance from Bayesian Perspective.} In conditional diffusion models, the conditional score function sampled from the posterior distribution $p(\mathbf{z}_t | c)$ can be decoupled by Bayes' theorem into a prior and a guidance term~\cite{dhariwal2021diffusion}:
\begin{equation}
\nabla_{\mathbf{z}_t} \log p(\mathbf{z}_t | c) = \nabla_{\mathbf{z}_t} \log p(\mathbf{z}_t) + \nabla_{\mathbf{z}_t} \log p(c | \mathbf{z}_t)
\end{equation}
To explicitly inject physical interventions, we model the constraint $c$ as a Boltzmann distribution governed by an energy function $\mathcal{E}$, such that $p(c | \mathbf{z}_t) \propto \exp(-\mathcal{E}(\mathbf{z}_t))$. Evaluating the logarithmic gradient of this term and substituting it into the Bayesian formulation yields the modified score function (as illustrated at the bottom of the Figure~\ref{fig:framework}):
\begin{equation}
\tilde{s}_\theta(\mathbf{z}_t) = s_\theta(\mathbf{z}_t) - \gamma \nabla_{\mathbf{z}_t} \mathcal{E}(\mathbf{z}_t)
\end{equation}
where $\gamma$ is the guidance scale. This mechanism demonstrates that subtracting the energy gradient pushes the denoising trajectory towards local minima that conform to defined constraints.

\textbf{Construction of Energy Landscapes.} To address the issue of spatial crowding, we define the \textbf{Spatial landscape} $\mathcal{E}_{clash}$ on the reconstructed coordinates $\hat{\mathbf{X_{sc}}}$, imposing a van der Waals repulsive penalty on non-adjacent atom pairs with distances below the threshold $\tau$:
\begin{equation}
\mathcal{E}_{clash} = \frac{1}{2} \sum_{a, b \notin \text{adjacent}} \max(0, \tau - D_{a,b})^2
\end{equation}
Simultaneously, given paired indices $(i, j) \in \mathcal{E}_{pair}$ from the target 2D graph, the \textbf{Topological landscape} $\mathcal{E}_{ss}$ is defined as the negative log-likelihood based on the valid pair set $\mathcal{V}$ to ensure adherence to base-pairing rules:
\begin{equation}
\mathcal{E}_{ss} = - \frac{1}{|\mathcal{E}_{pair}|} \sum_{(i,j) \in \mathcal{E}_{pair}} \log \Big( \sum_{(A,B) \in \mathcal{V}} \hat{\mathbf{P}}_{i, A} \hat{\mathbf{P}}_{j, B} + \epsilon \Big)
\end{equation}
where $\hat{\mathbf{P}}_{i, A}$ is the decoded probability of nucleotide $A$ at position $i$. At each denoising step $t$, we compute the \textbf{Joint landscape} $\mathcal{E}_{total} = \omega_{ss}\mathcal{E}_{ss} + \omega_{clash}\mathcal{E}_{clash}$ and backpropagate the gradients. This mechanism  guides the model to simultaneously fine-tune both the soft sequence probabilities and the $\chi$ angles, thereby achieving zero-shot generalization to physical constraints. For a detailed explanation of GeoVAE, latent diffusion, and the dual-guided sampling process, including complete algorithmic pseudocode and specific model hyperparameter configurations, please refer to Appendix~\ref{app:algorithms}.

\section{Experiments}
\label{experiments}
\subsection{Experimental Setup}
\label{experiment_setup}
\textbf{Datasets.} We construct the \textbf{GOLD Benchmark}, a high-quality, non-redundant RNA inverse folding dataset sourced from the PDB. After rigorous dual-stage clustering to prevent data leakage, the final dataset contains $7,583$ training, $1,144$ validation, and $598$ testing samples. For out-of-distribution (OOD) evaluation on highly complex targets, we incorporate the RNA subset from the \textbf{CASP15} challenge ~\citep{Das2023.04.25.53833}. Detailed construction protocols are provided in Appendix~\ref{app:dataset_construction}.

\textbf{Baselines.} We benchmark GOLD against state-of-the-art autoregressive and diffusion models for inverse folding: PiFold \cite{gao2023pifold}, RDesign \cite{tan2024rdesign}, gRNAde \cite{joshi2025grnade}, RiboDiffusion \cite{han2024ribodiffusion}, and SOLD \cite{si2026structure}. Baseline implementation specifics are detailed in Appendix~\ref{app:baseline_implementations}.

\textbf{Evaluation Metrics.} To ensure fair comparisons against 1D-only baselines, we employ external oracles to evaluate 1D \textbf{Sequence Recovery}, 2D Secondary Structure (\textbf{SS}) Recovery (via RNAfold \cite{hofacker1994fast}), and 3D structural \textbf{RMSD} (via AlphaFold3 \cite{abramson2024accurate} and RhoFold \cite{shen2024accurate}). Furthermore, to  validate the physical of GOLD's \textbf{native} geometry without relying on black-box refolding tools, we define two intrinsic metrics for ablation studies: \textbf{SS Validity} (topological pairing success rate) and \textbf{Clash Score} (steric collisions per 1,000 atoms). All formal mathematical definitions are detailed in Appendix~\ref{app:evaluation_metrics}.

\subsection{Main Results}
\label{main_results}

\begin{table}[ht]
    \centering
    \caption{Performance comparison of 3D RNA inverse folding methods. Evaluation is conducted on the internal GOLD Benchmark and the external CASP15 targets. Results are reported as mean $\pm$ standard deviation over five random seeds.}
    \label{tab:main_results}
    
    \small 
    \setlength{\tabcolsep}{4pt} 
    
    \begin{tabular}{l cccc}
        \toprule
        \multirow{2}{*}{\textbf{Method}} & \multirow{2}{*}{\textbf{Seq Recovery ($\uparrow$)}} & \multirow{2}{*}{\textbf{SS Recovery ($\uparrow$)}} & \multicolumn{2}{c}{\textbf{RMSD (\AA, $\downarrow$)}} \\
        \cmidrule(lr){4-5}
        & & & \textbf{RhoFold} & \textbf{AlphaFold3} \\
        \midrule
        
        \multicolumn{5}{c}{\cellcolor{gray!15}\textbf{GOLD Benchmark (In-Distribution)}} \\
        \midrule
        PiFold         & $0.4123 \pm 0.0354$ & $0.6184 \pm 0.0648$ & $17.26 \pm 3.56$ & $18.73 \pm 4.27$ \\
        RDesign        & $0.4430 \pm 0.0326$ & $0.6751 \pm 0.1106$ & $14.65 \pm 4.36$ & $15.12 \pm 3.85$ \\
        gRNAde         & $0.5453 \pm 0.0873$ & $0.6450 \pm 0.1183$ & $15.38 \pm 4.84$ & $16.10 \pm 4.37$ \\
        RiboDiffusion  & $0.5148 \pm 0.0574$ & $0.6948 \pm 0.0846$ & $13.07 \pm 3.19$ & $13.89 \pm 3.92$ \\
        SOLD           & $\underline{0.5843 \pm 0.0635}$ & $\underline{0.7366 \pm 0.0983}$ & $\underline{12.61 \pm 2.79}$ & $\underline{13.65 \pm 4.06}$ \\
        \midrule
        \textbf{GOLD (Base)} & $0.6249 \pm 0.0283$ & $0.7785 \pm 0.0782$ & $11.99 \pm 2.84$ & $13.26 \pm 4.34$ \\
        \textbf{GOLD (Dual-Guided)} & $\mathbf{0.6380 \pm 0.0292}$ & $\mathbf{0.8186 \pm 0.0724}$ & $\mathbf{10.65 \pm 2.63}$ & $\mathbf{12.31 \pm 3.63}$ \\
        
        \midrule
        \midrule
        
        \multicolumn{5}{c}{\cellcolor{gray!15}\textbf{CASP15 (Out-of-Distribution)}} \\
        \midrule
        PiFold         & $0.3156 \pm 0.0362$ & $0.3407 \pm 0.0564$ & $36.23 \pm 4.94$ & $35.43 \pm 6.72$ \\
        RDesign        & $0.3640 \pm 0.0305$ & $0.4250 \pm 0.0568$ & $33.18 \pm 4.86$ & $38.42 \pm 5.86$ \\
        gRNAde         & $0.5161 \pm 0.0204$ & $0.4266 \pm 0.0478$ & $33.89 \pm 5.53$ & $39.23 \pm 5.39$ \\
        RiboDiffusion  & $0.4959 \pm 0.0265$ & $0.6134 \pm 0.0505$ & $31.14 \pm 3.57$ & $30.74 \pm 4.28$ \\
        SOLD           & $\underline{0.5877 \pm 0.0106}$ & $\underline{0.6367 \pm 0.0481}$ & $\underline{28.87 \pm 3.23}$ & $\underline{27.74 \pm 4.04}$ \\
        \midrule
        \textbf{GOLD (Base)} & $0.6187 \pm 0.0246$ & $0.6521 \pm 0.0484$ & $28.32 \pm 3.76$ & $27.17 \pm 4.74$ \\
        \textbf{GOLD (Dual-Guided)} & $\mathbf{0.6624 \pm 0.0308}$ & $\mathbf{0.7089 \pm 0.0427}$ & $\mathbf{27.64 \pm 3.57}$ & $\mathbf{26.18 \pm 3.93}$ \\
        \bottomrule
    \end{tabular}
\end{table}

We evaluate GOLD against several representative methods on both our internal GOLD Benchmark and the challenging out-of-distribution CASP15 targets. As shown in Table~\ref{tab:main_results}, our full model (\textbf{GOLD (Dual-Guided)}), consistently achieves the best performance in all evaluation metrics. Specifically, it demonstrates robust improvements in 1D sequence recovery and 2D topological accuracy, while yielding more physically feasible 3D structures with lower RMSD scores (as evaluated by both RhoFold and AlphaFold3). Notably, even in the absence of explicit guidance, \textbf{GOLD (Base)} remains highly competitive and consistently outperforms the strongest baseline (SOLD). This result further underscores the inherent effectiveness of our unified sequence-structure latent space in capturing critical structural dependencies.

\textbf{Impact of the Unified Sequence-Structure Manifold.} As observed in Table~\ref{tab:main_results}, the better performance of SOLD compared to RiboDiffusion highlights the benefits of projecting discrete sequences into a continuous latent space. Moreover, our unguided baseline \textbf{GOLD (Base)}, achieves further improvements, outperforming SOLD across all metrics on the GOLD Benchmark (e.g., sequence recovery of 0.6249 vs. 0.5843). We attribute this performance gain primarily to the difference in representation spaces. SOLD relies on a unimodal sequence latent space, whereas GOLD further introduces the \textbf{joint modeling of sequences and side-chain angles}. By encoding discrete sequence tokens and continuous torsion angles into an $SO(2)$-equivariant manifold, GOLD provides an optimization landscape for diffusion models that is inherently capable of capturing 3D physical dependencies.

\textbf{Impact of Dual-Guided Sampling.} Although the unified latent space provides a strong structural prior, relying solely on this latent representation may not fully resolve fine-grained physical constraints. To mitigate this issue, we introduce a dual-guided mechanism into the sampling process. In our experiments, we set $\omega_{ss} = 1.0$ to enforce topological pairing and $\omega_{clash} = 0.1$ to balance spatial constraints. As shown in Table~\ref{tab:main_results}, on the GOLD benchmark, this dual-guided mechanism improves the 2D topological pairing accuracy (SS Recovery) from $0.7785$ to $0.8186$ over the unguided baseline, while further reducing the AlphaFold3-based RMSD from $13.26$ \AA\ to $12.31$ \AA. Unlike methods that require further fine-tuning, this dual-guided mechanism is completely training-free.

\textbf{Generalization on CASP15 Targets.} The CASP15 dataset provides a out-of-distribution (OOD) test involving highly complex, unseen RNA conformations. As Table~\ref{tab:main_results} shows, even without explicit guidance, GOLD (Base) outperforms the strongest baseline (SOLD) in all metrics. Performance can be further enhanced by employing dual-guided sampling (GOLD), resulting in optimal overall performance, including a sequence recovery rate of 0.6624 and an AlphaFold3 RMSD of 26.18Å. Ultimately, these results demonstrate that GOLD performs strongly in zero-shot generalization: the joint sequence and side-chain angle modeling provides a strong foundational advantage on novel topologies and dual-guided sampling further enhances this advantage.

\subsection{Ablation Studies}
\label{ablations}

To systematically analyze the contributions of key components in GOLD, we conduct further ablation experiments on the internal GOLD Benchmark dataset.

\textbf{Explicit Sequence--Torsion Modeling.} Unlike conventional RNA inverse-folding methods that primarily use geometric features as structural conditions, GOLD explicitly models the glycosidic torsion angle $\chi$ alongside nucleotide identity in a unified latent representation. To evaluate the contribution of explicit side-chain torsion modeling, we introduce \textbf{GOLD (Seq-only)}, a controlled variant that removes the $SO(2)$-aware fusion module and the geometric losses $\mathcal{L}_{\mathrm{angle}}$ and $\mathcal{L}_{\mathrm{coord}}$. Unlike SOLD, which also performs sequence-only latent diffusion but employs a different VAE, GOLD (Seq-only) retains the same latent dimension, RNA-FM alignment, backbone-conditioned GVP, diffusion backbone, and training settings as GOLD (Base). Both variants are evaluated without gradient guidance to isolate the effect of joint sequence--torsion modeling. Importantly, GOLD (Seq-only) still incorporates backbone geometry as structural conditioning, allowing us to distinguish explicit torsion modeling from the conventional use of geometric features. As shown in Table~\ref{tab:seq_only_ablation}, GOLD (Seq-only) performs comparably to SOLD across sequence recovery, secondary-structure recovery, and RhoFold-based 3D evaluation, whereas GOLD (Base) consistently outperforms both sequence-only variants. These results demonstrate that the improvements of GOLD are not solely attributable to latent diffusion or foundation-model alignment, but benefit from explicitly coupling sequence identity with side-chain torsion geometry in a unified latent representation.

\begin{table}[ht]
    \centering
    \caption{Ablation of explicit sequence--torsion modeling on the GOLD Benchmark. All methods are evaluated without gradient guidance.}
    \label{tab:seq_only_ablation}
    \small
    \setlength{\tabcolsep}{5pt}
    \resizebox{\linewidth}{!}{
    \begin{tabular}{lcccc}
        \toprule
        \textbf{Method}
        & \textbf{Torsion $\chi$}
        & \textbf{Seq Recovery ($\uparrow$)}
        & \textbf{SS Recovery ($\uparrow$)}
        & \textbf{RhoFold RMSD (\AA, $\downarrow$)} \\
        \midrule
        SOLD
        & $\times$
        & $0.5843 \pm 0.0635$
        & $0.7366 \pm 0.0983$
        & $12.61 \pm 2.79$ \\
        GOLD (Seq-only)
        & $\times$
        & $0.5897 \pm 0.0346$
        & $0.7402 \pm 0.1036$
        & $12.74 \pm 2.93$ \\
        \textbf{GOLD (Base)}
        & $\checkmark$
        & $\mathbf{0.6249 \pm 0.0283}$
        & $\mathbf{0.7785 \pm 0.0782}$
        & $\mathbf{11.99 \pm 2.84}$ \\
        \bottomrule
    \end{tabular}}
\end{table}

\textbf{Impact of Geometric Supervision.} To further investigate the contribution of geometric supervision in GeoVAE, we separately remove the angle loss $\mathcal{L}_{\mathrm{angle}}$ and coordinate loss $\mathcal{L}_{\mathrm{coord}}$, while keeping the remaining architecture and training settings unchanged. As shown in Table~\ref{tab:geometric_loss_ablation}, removing $\mathcal{L}_{\mathrm{angle}}$ mainly degrades torsion prediction accuracy, while $\mathcal{L}_{\mathrm{coord}}$ still provides indirect geometric supervision. In contrast, removing $\mathcal{L}_{\mathrm{coord}}$ substantially increases coordinate reconstruction error and KAE, despite maintaining relatively accurate torsion predictions. These results demonstrate that the two geometric losses play complementary roles, and accurate torsion prediction alone is insufficient to ensure geometrically consistent 3D reconstruction.

\begin{table}[ht]
    \centering
    \caption{Ablation of geometric supervision losses in GeoVAE on the GOLD Benchmark.}
    \label{tab:geometric_loss_ablation}
    \small
    \setlength{\tabcolsep}{6pt}
    \begin{tabular}{lccc}
        \toprule
        \textbf{Variant}
        & \textbf{Angle MSE ($\downarrow$)}
        & \textbf{Coord MSE ($\downarrow$)}
        & \textbf{KAE (\AA, $\downarrow$)} \\
        \midrule
        w/o $\mathcal{L}_{\mathrm{angle}}$
        & $0.0546$ & $0.0388$ & $0.2802$ \\
        w/o $\mathcal{L}_{\mathrm{coord}}$
        & $0.0136$ & $0.1611$ & $0.6812$ \\
        \textbf{Full GeoVAE}
        & $\mathbf{0.0109}$
        & $\mathbf{0.0287}$
        & $\mathbf{0.2466}$ \\
        \bottomrule
    \end{tabular}
\end{table}

\textbf{Impact of Alignment on Foundation Model Representation.} In Table~\ref{tab:ablation_fm}, we evaluate the impact of Foundation Model (FM) alignment by comparing a variant without an FM (w/o FM) to those incorporating a generalist genomic model (Evo2~\cite{Brixi2026}) or an RNA-specific model (RNA-FM~\cite{chen2022interpretable}). The results show that while the underlying $SO(2)$ geometric manifold provides a robust structural baseline, integrating FM alignment consistently improves metrics for Sequence and Secondary Structure Recovery. We attribute this to the co-evolutionary priors embedded in large language models, which provide sequence-level constraints that help guide the generation process to produce biologically plausible sequences. Furthermore, the model using domain-specific RNA-FM demonstrated slightly better but more consistent performance compared to the generalist Evo2. This suggests that, for RNA inverse folding, evolutionary signals derived from RNA databases may provide more relevant semantic context than broad genomic representations.

\begin{table}[ht]
    \centering
    \caption{Ablation study on Foundation Model (FM) representation alignment.}
    \label{tab:ablation_fm}
    \small
    \setlength{\tabcolsep}{6pt} 
    \begin{tabular}{l cccc}
        \toprule
        \multirow{2}{*}{\textbf{Model Variant}} & \multirow{2}{*}{\textbf{Seq Recovery ($\uparrow$)}} & \multirow{2}{*}{\textbf{SS Recovery ($\uparrow$)}} & \multicolumn{2}{c}{\textbf{RMSD (\AA, $\downarrow$)}} \\
        \cmidrule(lr){4-5}
        & & & \textbf{RhoFold} & \textbf{AlphaFold3} \\
        \midrule
        GOLD (w/o FM)             & $0.6217$ & $0.7792$ & $11.23$ & $12.58$ \\
        GOLD (w/ Evo2)            & $0.6302$ & $0.7922$ & $10.85$ & $12.40$ \\
        \textbf{GOLD (w/ RNA-FM)} & $\mathbf{0.6380}$ & $\mathbf{0.8186}$ & $\mathbf{10.65}$ & $\mathbf{12.31}$ \\
        \bottomrule
    \end{tabular}
\end{table}

\textbf{Dual-Guided Sampling.} To distinguish the effects of the guidance gradients from the external folding tools, we evaluate the SS Validity and Clash Score directly on the model's native outputs (Table~\ref{tab:ablation_guide}). While the baseline model exhibits noticeable side-chain clashes in the absence of explicit guidance, the introduction of topological guidance ($\omega_{ss}$) and spatial guidance ($\omega_{clash}$) enhances Watson-Crick pairing validity and suppresses side-chain collisions. However, relying heavily on a single constraint (e.g., $\omega_{clash}=1.0$) can distort the latent manifold, leading to a decrease in Sequence Recovery metric. By balancing both constraints ($\omega_{ss}=1.0, \omega_{clash}=0.1$), the dual-guided sampling setup achieves a favorable trade-off. It not only improves Sequence Recovery metric but also ensures the validity of the topology and minimizes spatial conflicts, indicating that incorporating these physical gradients can effectively guide the generation process without compromising overall sequence quality.

\begin{table}[ht]
    \centering
    \caption{Ablation study on the training-free Guidance mechanism.}
    \label{tab:ablation_guide}
    \small
    \setlength{\tabcolsep}{6pt}
    \begin{tabular}{lcc ccc}
        \toprule
        \multirow{2}{*}{\textbf{Guidance Setup}} & \multicolumn{2}{c}{\textbf{Weights}} & \multirow{2}{*}{\textbf{Seq Recovery ($\uparrow$)}} & \multirow{2}{*}{\textbf{SS Validity ($\uparrow$)}} & \multirow{2}{*}{\textbf{Clash Score ($\downarrow$)}} \\
        \cmidrule(lr){2-3}
        & $\omega_{ss}$ & $\omega_{clash}$ & & & \\
        \midrule
        Base (No Guidance) & $0.0$ & $0.0$  & $0.6249$ & $0.9040$ & $4.51$ \\
        \midrule
        \multirow{3}{*}{SS Guidance Only} 
        & $1.0$ & $0.0$ & $0.6267$ & $0.9343$ & $4.32$ \\
        & $2.0$ & $0.0$ & $0.6264$ & $0.9356$ & $4.43$ \\
        & $3.0$ & $0.0$ & $0.6268$ & $\mathbf{0.9382}$ & $4.38$ \\
        \midrule
        \multirow{4}{*}{Clash Guidance Only} 
        & $0.0$ & $0.05$ & $0.6362$ & $0.9012$ & $3.11$ \\
        & $0.0$ & $0.1$  & $0.6378$ & $0.9028$ & $2.76$ \\
        & $0.0$ & $0.5$  & $0.6353$ & $0.9061$ & $\mathbf{2.28}$ \\
        & $0.0$ & $1.0$  & $0.6269$ & $0.8938$ & $2.77$ \\
        \midrule
        \textbf{Dual-Guidance} & $\mathbf{1.0}$ & $\mathbf{0.1}$ & $\mathbf{0.6380}$ & $0.9290$ & $2.67$ \\
        \bottomrule
    \end{tabular}
\end{table}

Additional appendices detail our evaluation results and mechanism analysis. 
Specifically, we evaluate the advantages of the unified latent manifold and compare our dynamic gradient guidance with direct feature utilization. We further assess the one-to-many nature of RNA inverse folding through sequence novelty, structural success rate, and intra-target diversity. Additionally, we investigate the interaction and sensitivity of the two physical guidance objectives through a two-dimensional weight sweep and KL-based analysis of the guided reverse process (Appendix~\ref{app:extended_ablations}).
Furthermore, we validate the effectiveness of the $SO(2)$-equivariant layer by comparing it with alternative sequence--torsion fusion strategies, including Trig-Concat and Fourier--FiLM (Appendix~\ref{app:so2_rationale}), and we also analyze latent dimensionality impacts (Appendix~\ref{app:latent_dim_tradeoff}), and evaluate scalability on long RNA targets (Appendix~\ref{app:long_sequence}).

\subsection{Case Studies}
\label{main_case_study}
To qualitatively evaluate GOLD, we analyze the structural validity of the model's native outputs and use external tools to assess the actual foldability of the generated sequences.

\begin{figure}[htbp]
    \centering
    \includegraphics[width=1.0\textwidth]{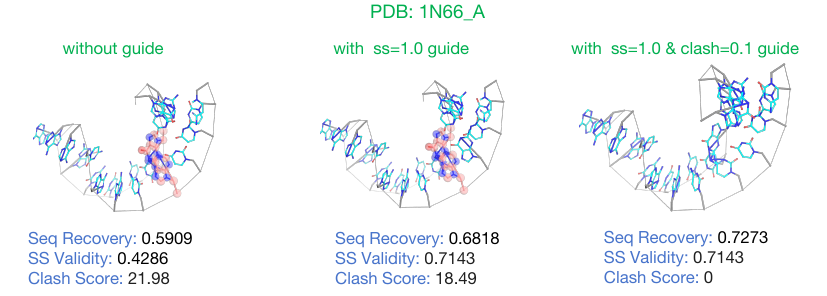} 
    \caption{\textbf{Native structure reconstructions.} \textbf{Left}: Unguided (side-chain clashes). \textbf{Middle}: SS-only guided (improved pairing, but residual clashes). \textbf{Right}: Dual-guided (collisions resolved).}
    \label{fig:case_internal}
\end{figure}

\textbf{Native Structural Validity.} Figure~\ref{fig:case_internal} visualizes native local reconstructions without external relaxation. Although the unguided baseline exhibits significant atomic overlap, and the model using SS-only guidance improves base pairing but still exhibits residual steric conflicts (red spheres), dual guidance eliminates all collisions while maintaining pairing fidelity. This demonstrates that joint constraints are essential for local structural plausibility (more details see Appendix~\ref{app:guidance_dynamics}).

\begin{figure}[htbp]
    \centering
    \includegraphics[width=1.0\textwidth]{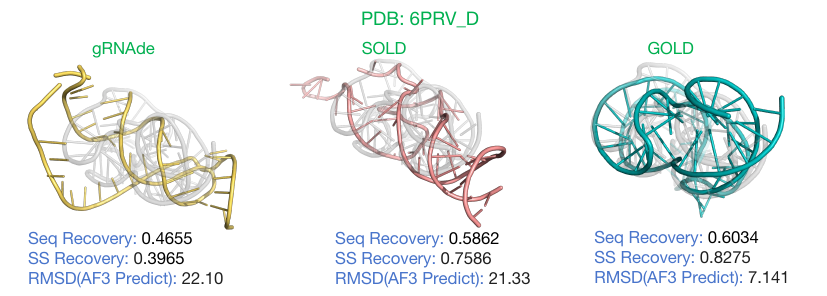} 
    \caption{\textbf{AlphaFold3 folding validation for target \texttt{6PRV\_D}.} Compared to baselines, the structure designed by GOLD aligns closely with the native ground truth (gray).}
    \label{fig:case_external}
\end{figure}

\textbf{External Folding Validation.} We assess foldability using AlphaFold3 predictions (Figure ~\ref{fig:case_external} for target \textbf{6PRV\_D}). Baseline models (gRNAde, SOLD) exhibit spatial discrepancies compared to the actual structure, whereas the sequences designed by GOLD show a high degree of alignment with the native backbone. This confirms that our method can effectively translate geometry information into viable sequences (see more in Appendix~\ref{app:extended_case_study}).

\section{Conclusion}
\label{conclusion}
In this work, we propose GOLD, a geometry-aware latent diffusion framework for RNA inverse folding. By mapping discrete sequences and continuous side-chain geometries into a unified $SO(2)$-equivariant latent space, GOLD effectively addresses the modality gap in biomolecular representation. Furthermore, our dual-guided mechanism explicitly incorporates physical gradients into the sampling process. This strategy reduces steric clashes while enhancing the topological pairing validity without relying on complex reinforcement learning fine-tuning. Evaluations on internal benchmarks and CASP15 targets demonstrate GOLD exhibits robust structural fidelity and generalization capabilities.

\textbf{Limitations and Future Work.} 
First, calculating physical gradients during the sampling process increases inference time. Future work could explore distillation techniques to directly learn these physical rules, thereby achieving faster generation speeds. Second, while GOLD currently focuses on the standard fixed-backbone inverse folding paradigm, extending our latent manifold to support flexible-backbone co-design or multi-state conformational sampling remains a promising direction.

\section*{Acknowledgments}
\label{acknowledgments}
This work was supported by the National Natural Science Foundation of China (Nos. 82394432, 92249302), the Shanghai Municipal Science and Technology Major Project (No. 2023SHZDZX02), and the AI for Science Program of the Shanghai Municipal Commission of Economy and Information (No. 2025-GZL-RGZN-BTBX-01008). We also acknowledge the support of the Novalnspire platform (Shanghai Academy of Artificial Intelligence for Science) and the CFFF platform (Fudan University) for providing computational resources.

\bibliographystyle{plainnat}
\bibliography{references}

\clearpage
\appendix



\thispagestyle{plain}

\begingroup
\small
\linespread{1.15}\selectfont
\setlength{\parindent}{0pt}
\setlength{\parskip}{0pt}


\newcommand{\apptocsection}[3]{%
    \noindent
    \textbf{\hyperref[#1]{#2\quad #3}}%
    \nobreak\dotfill\pageref{#1}\par
    \vspace{0.3em}
}

\newcommand{\apptocsubsection}[3]{%
    \noindent
    \hspace*{1.5em}%
    \hyperref[#1]{#2\quad #3}%
    \nobreak\dotfill\pageref{#1}\par
    \vspace{0.12em}
}

\newcommand{\apptocgap}{%
    \par\vspace{0.5em}\vfill
}

\vspace*{1.5em}
\begin{center}
    {\LARGE \textbf{Appendix}}\\[1.2em]
    {\Large \textbf{Table of Contents}}
\end{center}
\vspace{1.5em}

\apptocsection
    {app:motivation}
    {A}
    {Design Principles}

\apptocgap

\apptocsection
    {app:algorithms}
    {B}
    {Algorithm Details}

\apptocsubsection
    {app:geovae_algo}
    {B.1}
    {GeoVAE Details}

\apptocsubsection
    {app:guided_sampling}
    {B.2}
    {Latent Diffusion: Score-Matching Training and Dual-Guided Sampling}

\apptocsubsection
    {app:training_details}
    {B.3}
    {Training Details}

\apptocgap

\apptocsection
    {app:experimental_setup}
    {C}
    {Experimental Details}

\apptocsubsection
    {app:dataset_construction}
    {C.1}
    {Dataset Construction}

\apptocsubsection
    {app:baseline_implementations}
    {C.2}
    {Baseline Implementations}

\apptocsubsection
    {app:evaluation_metrics}
    {C.3}
    {Evaluation Metrics}

\apptocgap

\apptocsection
    {app:long_sequence}
    {D}
    {Performance and Cost Analysis Across Sequence Lengths}

\apptocsubsection
    {app:long_rna}
    {D.1}
    {Performance of Long RNA}

\apptocsubsection
    {app:inference_efficiency}
    {D.2}
    {Inference Efficiency Across Varying Lengths}

\apptocgap

\apptocsection
    {app:so2_rationale}
    {E}
    {Why SO(2) Layer is Necessary in GeoVAE?}

\apptocsubsection
    {app:so2_poc}
    {E.1}
    {Experimental Setup}

\apptocsubsection
    {app:so2_observation_1}
    {E.2}
    {Mitigating Modality Collapse}

\apptocsubsection
    {app:so2_observation_2}
    {E.3}
    {Structural Optimization Dynamics}

\apptocsubsection
    {app:so2_fusion_ablation}
    {E.4}
    {Quantitative Comparison of Sequence--Torsion Fusion Strategies}

\apptocgap

\apptocsection
    {app:latent_dim_tradeoff}
    {F}
    {Discussion on Latent Dimension}

\apptocsubsection
    {app:reconstruction_vs_generation}
    {F.1}
    {Decoupling Reconstruction Fidelity from Generative Capacity}

\apptocsubsection
    {app:explain_latent_dim_robustness}
    {F.2}
    {Robustness to Latent Dimension Scaling}

\apptocgap

\apptocsection
    {app:extended_ablations}
    {G}
    {Extended Ablation Studies and Generative Evaluation}

\apptocsubsection
    {app:raw_space_comparison}
    {G.1}
    {Unified Latent Manifold vs Raw-Space Co-Diffusion}

\apptocsubsection
    {app:ss_leakage_ablation}
    {G.2}
    {Input Feature vs Gradient Guidance}

\apptocsubsection
    {app:one_to_many}
    {G.3}
    {One-to-Many Generative Evaluation}

\apptocsubsection
    {app:guidance_sensitivity}
    {G.4}
    {Interaction and Sensitivity of Physical Guidance}

\apptocgap

\apptocsection
    {app:guidance_dynamics}
    {H}
    {Analysis of Dual-Guidance Dynamics}

\apptocsubsection
    {app:guidance_ss}
    {H.1}
    {Topological Guidance ($\omega_{ss}$)}

\apptocsubsection
    {app:clash_ablation}
    {H.2}
    {Clash Guidance ($\omega_{clash}$)}

\apptocgap

\apptocsection
    {app:extended_case_study}
    {I}
    {Extended Case Studies}

\apptocsubsection
    {app:extended_case_study_I}
    {I.1}
    {Case Study I}

\apptocsubsection
    {app:extended_case_study_II}
    {I.2}
    {Case Study II}

\vfill

\endgroup

\clearpage

\section{Design Principles}
\label{app:motivation}

\begin{figure}[h]
  \centering
  \includegraphics[width=0.8\linewidth]{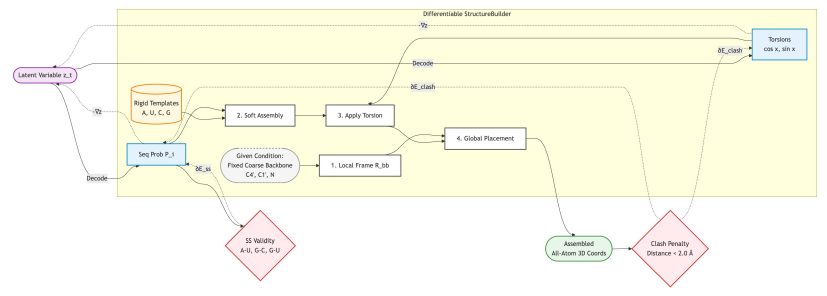} 
  \caption{\textbf{StructureBuilder and Gradient Flow.} The architecture serves as a differentiable routing board, allowing gradients from 2D topological and 3D physical constraints to steer the latent sampling process.}
  \label{fig:pipeline_appendix}
\end{figure}

\textbf{Motivation: Enhancing Physical Validity via Geometric Modeling.}  Conventional RNA inverse folding models often lack a mechanism to bridge 1D sequences and 3D physical space. To address this issue, GOLD implements the differentiable \textbf{StructureBuilder} as its core geometric engine. This module realizes the operator $\mathcal{B}$ defined in main text Section ~\ref{preliminary}. It utilizes sequence probabilities and side-chain angles to map latent predictions into 3D coordinates. This design serves a dual purpose. First, it couples discrete sequences with geometric angles at a physical level. This intrinsically anchors the joint generative distribution in a more physically plausible space. Second, it provides a fully differentiable pathway for direct physical energy feedback. This enables the model to explicitly resolve residual spatial conflicts.

\textbf{Construction and Empirical Validation of the StructureBuilder.} To map predicted latent variables back into physical space for spatial constraint perception, the StructureBuilder fuses sequence probabilities and side-chain angles with the fixed backbone via a four-stage differentiable cascade:
\begin{itemize}
     \item \textbf{Stage I: Soft Semantic Assembly.} Weighted interpolation of standard base templates using the predicted sequence probabilities $\mathbf{P}_i$.
     \item \textbf{Stage II: Torsional Application.} Converting the predicted $\mathrm{SO}(2)$ angle $\chi_i$ into a 3D rotation matrix $\mathbf{R}_{\chi_i}$ to locally rotate the soft-assembled side-chain.
    \item \textbf{Stage III: Conditional Frame Construction.} Deterministically deriving local reference frames $\mathbf{R}_{bb}^{(i)}$ from the fixed backbone atoms.
    \item \textbf{Stage IV: Global Placement.} Projecting the side-chains into global coordinates $\mathbf{X}_{final}^{(i)}$, providing the 3D coordinate basis for calculating physical collision energies and gradients.
\end{itemize}

The physical foundation of this assembly mechanism relies on the planar rigid-body assumption of RNA nucleobases. To validate this assumption, we constructed a standard template bank $\mathbf{X}_{template} \in \mathbb{R}^{4 \times 14 \times 3}$. Specifically, we extracted the full-atom nucleotide coordinates from the GOLD Benchmark dataset. After aligning these extracted coordinates to a unified local relative coordinate system, we averaged them by base identity (A, U, C, G) to obtain idealized standard templates. Subsequently, we validated on the extensive GOLD Training set ($\approx$ 8,000 PDB structures). Restricting the internal DoF to the 1D rotation $\chi$ yields a \textbf{KAE} below $0.2$\text{\AA}. This result confirms that $\chi$ effectively captures side-chain geometry, which validates our strategy of using $\chi$ as the primary intermediate geometric variable.

\textbf{Gradient Flow of the Physical Guidance Mechanism.} 
The \textbf{StructureBuilder} acts as a bridge between physical signals and the sequence space during the sampling process (see Figure~\ref{fig:pipeline_appendix}). At each timestep $t$, latent variables are decoded into sequence probabilities and side-chain angles, which are then assembled into coordinates. This supports two guidance pathways:
\begin{itemize}
    \item \textbf{Pathway 1: Topological Guidance.} Calculating pairing probabilities to refine sequences.
    \item \textbf{Pathway 2: Physical Guidance.} Calculating clash energies between side-chains. The derived analytical gradients are decomposed to simultaneously update the side-chain orientations and sequence identities. Specifically:
    \begin{enumerate}
        \item To the side-chain torsion angle: Adjusting orientations to avoid collisions.
        \item To the sequence probability: Modifying the base composition to reduce the probability of physical conflicts, resulting in more physically plausible sequences.
    \end{enumerate}
\end{itemize}

\section{Algorithm Details}
\label{app:algorithms}

\subsection{GeoVAE Details}
\label{app:geovae_algo}
The GeoVAE is tasked with the objective of compressing discrete 1D nucleic acid sequences and continuous 3D spatial geometries into a smooth, compact, low-dimensional continuous manifold. To ensure that the latent space possesses both "geometric optimal" and "biological semantics", we incorporated SO(2) Rotation-Aware Layers into the forward pass to process torsion angles, and introduced a "Foundation Model Representation Alignment" mechanism to distill large-scale pre-trained semantic information (e.g., from RNA-FM) into the latent space.

The complete training procedure is detailed in Algorithm~\ref{alg:geovae}. During the encoding phase, the model outputs the posterior distribution parameters $\mu$ and $\log\sigma^2$. Concurrently, it projects the externally provided LLM sequence embeddings $H_{LLM}$ into the latent space (yielding $z_{teacher}$), and employs an alignment loss ($\mathcal{L}_{repa}$) to constrain the VAE's distribution center $\mu$ to a desired manifold. During the decoding phase, the model first reconstructs the sequence logits and continuous angles. Subsequently, by utilizing the operator $\mathcal{B}$, it recovers the spatial coordinates $\hat{X}_{sc}$ of side-chain conditioned on the coarse backbone $X_{bb}$ while maintaining the connectivity of computational graph. The multi-task loss function is scaled by a set of hyperparameters and optimized jointly end-to-end.

\begin{algorithm}[ht]
\caption{GeoVAE Training Forward Pass and Loss Computation}
\label{alg:geovae}
\begin{algorithmic}[1]
\REQUIRE Graph batch $\mathcal{G} = \{S, \chi, X_{bb}, X_{sc}\}$, LLM embeddings $H_{LLM}$, Loss weights $\mathbf{w}$
\ENSURE Updated Encoder ($\phi$), Decoder ($\theta$), and Projection ($\psi$) parameters
\STATE \textbf{// 1. SO(2)-Aware Encoding Phase}
\STATE $E_{seq} \gets \text{Embedding}(S)$
\STATE $H_{enc} \gets \text{RotationAwareEncode}_\phi(E_{seq}, \chi)$ \hfill \textit{$\triangleright$ Fuses seq and angles in $SO(2)$ domain}
\STATE $\mu, \log\sigma^2 \gets \text{GraphEncoder}_\phi(H_{enc}, X_{coarse})$
\STATE \textbf{// 2. Latent Reparameterization \& LLM Alignment}
\STATE $\epsilon \sim \mathcal{N}(0, I)$
\STATE $z \gets \mu + \epsilon \odot \exp(\frac{1}{2} \log\sigma^2)$
\STATE $z_{teacher} \gets \text{LatentProjection}_\psi(H_{LLM})$ \hfill \textit{$\triangleright$ Distill RNA-FM semantics}
\STATE $\mathcal{L}_{repa} \gets 1 - \cos(\mu, z_{teacher})$
\STATE \textbf{// 3. Decoding Phase}
\STATE $H_{dec} \gets \text{GraphDecoder}_\theta(z)$
\STATE $\hat{S}_{logits}, \hat{\chi} \gets \text{RotationAwareDecode}_\theta(H_{dec})$
\STATE \textbf{// 4. Differentiable Structure Building}
\STATE $\hat{P}_{seq} \gets \text{Softmax}(\hat{S}_{logits})$ \hfill \textit{$\triangleright$ Differentiable relaxation}
\STATE $\hat{X}_{sc} \gets \text{StructureBuilder}(\hat{P}_{seq}, \hat{\chi}, X_{bb})$ \hfill \textit{$\triangleright$ Recovers 3D coordinates}
\STATE \textbf{// 5. Multi-task Loss Computation}
\STATE $\mathcal{L}_{seq} \gets \text{CrossEntropy}(\hat{S}_{logits}, S)$
\STATE $\mathcal{L}_{angle} \gets \text{MSE}(\hat{\chi}, \chi)$
\STATE $\mathcal{L}_{coord} \gets \text{MSE}(\hat{X}_{sc}, X_{sc})$
\STATE $\mathcal{L}_{KL} \gets -\frac{1}{2} \sum (1 + \log\sigma^2 - \mu^2 - \exp(\log\sigma^2))$
\STATE $\mathcal{L}_{total} \gets w_{seq}\mathcal{L}_{seq} + w_{angle}\mathcal{L}_{angle} + w_{coord}\mathcal{L}_{coord} + w_{kl}\mathcal{L}_{KL} + w_{repa}\mathcal{L}_{repa}$
\STATE \textbf{Update} $\phi, \theta, \psi$ by descending $\nabla_{\phi, \theta, \psi} \mathcal{L}_{total}$
\end{algorithmic}
\end{algorithm}

\subsection{Latent Diffusion: Score-Matching Training and Dual-Guided Sampling}
\label{app:guided_sampling}

Our diffusion generative framework is composed of two core operational phases: continuous score-matching on the latent manifold (Training) and physics-aware reverse denoising (Inference). 

\textbf{Diffusion Training Phase.} 
As detailed in Algorithm~\ref{alg:diffusion_train}, we freeze the weights of the pre-trained GeoVAE, utilizing it as a high-fidelity latent tokenizer during training. For a given native macromolecular graph, we extract its latent representation $z_0$ through the GeoVAE encoder. We uniformly sample a timestep $t \sim \mathcal{U}(1, T)$ and inject Gaussian noise into $z_0$ according to a pre-defined cosine noise schedule, yielding the noisy state $z_t$. Our denoising backbone (DiT) is designed to directly predict the expectation of the clean latent variable $\hat{z}_0$ and the network parameters $\theta$ are optimized by minimizing the loss of MSE.

\begin{algorithm}[ht]
\caption{Latent Diffusion Training}
\label{alg:diffusion_train}
\begin{algorithmic}[1]
\REQUIRE Training graph batch $\mathcal{G} = \{S, \chi, X_{bb}\}$, Frozen GeoVAE Encoder ($\phi^*$), DiT model ($\theta$)
\ENSURE Updated DiT parameters $\theta$
\STATE \textbf{// 1. Extract Clean Latent Representation}
\STATE $\mu, \log\sigma^2 \gets {\phi^*}(S, \chi, X_{bb})$
\STATE $\epsilon_{reparameter} \sim \mathcal{N}(0, I)$
\STATE $z_0 \gets \mu + \epsilon_{reparameter} \odot \exp(\frac{1}{2} \log\sigma^2)$ \hfill \textit{$\triangleright$ Ground-truth clean latent}
\STATE \textbf{// 2. Forward Diffusion (Add Noise)}
\STATE $t \sim \mathcal{U}(1, T)$ \hfill \textit{$\triangleright$ Sample random timestep}
\STATE $\epsilon_{noise} \sim \mathcal{N}(0, I)$
\STATE $\bar{\alpha}_t \gets \text{CosineSchedule}(t)$
\STATE $z_t \gets \sqrt{\bar{\alpha}_t} z_0 + \sqrt{1 - \bar{\alpha}_t} \epsilon_{noise}$ \hfill \textit{$\triangleright$ Obtain noisy latent state}
\STATE \textbf{// 3. Denoising Prediction and Optimization}
\STATE $\hat{z}_0 \gets \text{DiT}_\theta(z_t, t, X_{bb})$ \hfill \textit{$\triangleright$ Model predicts clean $z_0$}
\STATE $\mathcal{L}_{latent} \gets \text{MSE}(\hat{z}_0, z_0)$ \hfill \textit{$\triangleright$ Score-matching objective}
\STATE \textbf{Update} $\theta$ by descending $\nabla_\theta \mathcal{L}_{latent}$
\end{algorithmic}
\end{algorithm}

\textbf{Dual-Guided Sampling Phase.} 
During inference, we propose a dual-guided sampling algorithm (Algorithm~\ref{alg:dual_guidance}). At each denoise timestep $t$, we leverage the frozen GeoVAE decoder to project the predicted $\hat{z}_0$ back into the 1D/3D biological space. This allows us to compute the topological pairing energy $\mathcal{E}_{ss}$ and spatial repulsion energy $\mathcal{E}_{clash}$ natively in the physical space. The aggregated energy gradients are then backpropagated to the denoised latent state $z_t$. To prevent excessive guidance from disrupting the smoothness of the manifold, we dynamically anneal the guidance scale ($\gamma_t = t / T$), imposing strong constraints in the early stages while maintaining micro-fidelity in the final stages.

\begin{algorithm}[ht]
\caption{Dual-Guided Latent Diffusion Sampling}
\label{alg:dual_guidance}
\begin{algorithmic}[1]
\REQUIRE Coarse backbone $X_{bb}$, Native SS graph $E_{ss}$, Total steps $T$, Guidance weights $\omega_{ss}, \omega_{clash}$
\ENSURE Sampled sequence $\hat{S}$ and fine-grained side-chain coordinates $\hat{X}_{sc}$
\STATE $z_T \sim \mathcal{N}(0, \mathbf{I})$
\FOR{$t = T$ \TO $1$}
    \STATE $\hat{z}_0 \gets \text{DiT}_\theta(z_t, t, X_{bb})$ \hfill \textit{$\triangleright$ Predict clean latent expectation}
    \IF{$\omega_{ss} > 0$ \OR $\omega_{clash} > 0$}
        \STATE \textbf{Require} gradient tracking for $z_t$
        \STATE \textbf{// 1. Cross-space Differentiable Projection}
        \STATE $\hat{S}_{logits}, \hat{\chi}, \hat{X}_{sc} \gets \text{GeoVAE\_Decoder}_{\theta^*}(\hat{z}_0, X_{bb})$ 
        \STATE $\hat{P}_{seq} \gets \text{Softmax}(\hat{S}_{logits})$
        \STATE \textbf{// 2. Physical Energy Computation}
        \STATE $\mathcal{E}_{ss} \gets \text{ComputePairingEnergy}(\hat{P}_{seq}, E_{ss})$ \hfill \textit{$\triangleright$ Evaluates Watson-Crick rules}
        \STATE $\mathcal{E}_{clash} \gets \text{ComputeClashEnergy}(\hat{X}_{sc}, \text{threshold}=2.0\text{\AA})$ 
        \STATE $\mathcal{E}_{total} \gets \omega_{ss}\mathcal{E}_{ss} + \omega_{clash}\mathcal{E}_{clash}$
        \STATE \textbf{// 3. Backpropagation and Dynamic Annealed Update}
        \STATE $g \gets \nabla_{z_t} \mathcal{E}_{total}$ \hfill \textit{$\triangleright$ Backpropagate energy gradient to latent}
        \STATE $\gamma_t \gets t / T$ \hfill \textit{$\triangleright$ Dynamic linear annealing scale}
        \STATE $z_t \gets z_t - \gamma_t \cdot g$ \hfill \textit{$\triangleright$ Apply physics-based guidance}
    \ENDIF
    \STATE $z_{t-1} \gets \text{Denoise\_Step}(z_t, \hat{z}_0, t)$ \hfill \textit{$\triangleright$ Standard Denoised manifold transition}
\ENDFOR
\STATE \textbf{// Final Decode}
\STATE $\hat{S}_{logits}, \hat{A}, \hat{X}_{sc} \gets \text{GeoVAE\_Decoder}_{\theta^*}(z_0, X_{bb})$
\STATE \textbf{Return} $\argmax(\hat{S}_{logits})$, $\hat{X}_{sc}$
\end{algorithmic}
\end{algorithm}

\begin{figure}[htbp]
    \centering
    \includegraphics[width=0.95\textwidth]{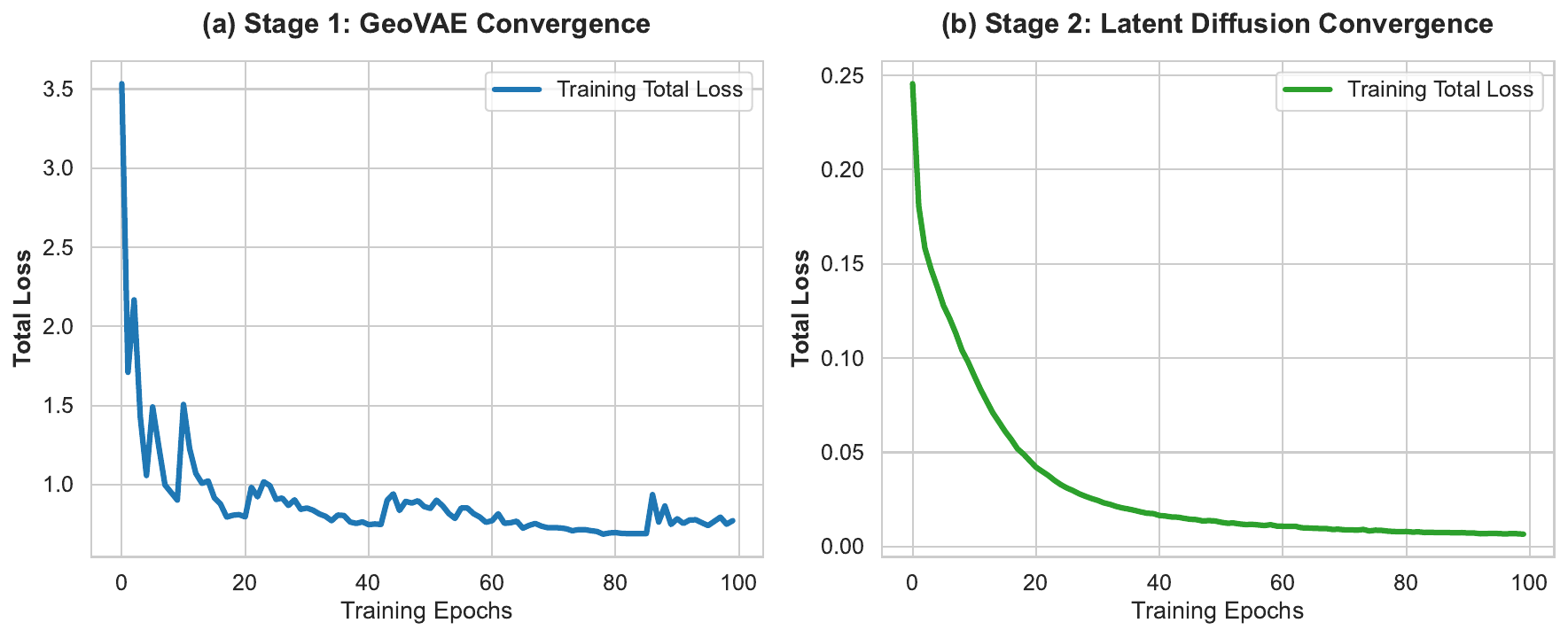} 
    \caption{\textbf{Convergence dynamics of the two-stage training process.} (a) Total validation loss of GeoVAE. (b) Total training loss of the Latent Diffusion stage.}
    \label{fig:training_dynamics}
\end{figure}

\subsection{Training Details}
\label{app:training_details}

To ensure reproducibility, we detail the hyperparameter configurations for our two-stage architecture in Table~\ref{tab:hyperparameters}.

\begin{table}[ht]
    \centering
    \caption{Detailed architectural configurations and training hyperparameters for the two-stage generative framework (GeoVAE and Latent Diffusion).}
    \label{tab:hyperparameters}
    \resizebox{0.95\textwidth}{!}{
    \begin{tabular}{lllc}
        \toprule
        \textbf{Training Stage} & \textbf{Module} & \textbf{Hyperparameter} & \textbf{Value} \\
        \midrule
        \multirow{13}{*}{\textbf{Stage 1: GeoVAE}} 
        & \multirow{6}{*}{\textit{Architecture}}
        & Node Hidden Dimension ($d_{model}$) & 256 \\
        && Latent Dimension ($d_{latent}$) & 8 \\
        && Encoder/Decoder Layers & 4 \\
        && Attention Heads & 4 \\
        && Dropout Rate & 0.1 \\
        && Pre-trained LLM Embed Dimension (RNA-FM) & 640 \\
        \cmidrule{2-4}
        & \multirow{7}{*}{\textit{Optimization}}
        & Base Learning Rate & $5 \times 10^{-4}$ \\
        && Weight Decay & $1 \times 10^{-4}$ \\
        && Batch Size (Per GPU) & 1 \\
        && Gradient Accumulation Steps & 16 \\
        && Training Epochs & 100 \\
        && Gradient Clipping & 1.0 \\
        && Loss Weights ($w_{coord}, w_{angle}, w_{seq}, w_{kl}$) & 10.0, 5.0, 1.0, 0.1 \\
        \midrule
        \multirow{15}{*}{\textbf{Stage 2: Latent Diffusion}} 
        & \multirow{4}{*}{\textit{GVP Condition}}
        & Node Hidden Dim (Scalar, Vector) & (128, 16) \\
        && Edge Hidden Dim (Scalar, Vector) & (32, 1) \\
        && GVP Layers & 4 \\
        && Spatial $k$-NN Graph & $k=10$ \\
        \cmidrule{2-4}
        & \multirow{4}{*}{\textit{DiT Backbone}}
        & Transformer Layers & 12 \\
        && Encoder Embedding Dimension & 128 \\
        && FFN Embedding Dimension & 512 \\
        && Attention Heads & 4 \\
        \cmidrule{2-4}
        & \multirow{7}{*}{\textit{Diffusion \& Optim}}
        & Diffusion Timesteps & 100 \\
        && Noise Schedule & Cosine \\
        && Base Learning Rate & $1 \times 10^{-4}$ \\
        && Weight Decay & $1 \times 10^{-4}$ \\
        && Batch Size & 32 \\
        && Training Epochs & 100 \\
        && Gradient Accumulation Steps & 1 \\
        \bottomrule
    \end{tabular}
    }
\end{table}

\textbf{Stage 1: GeoVAE Training.}
The GeoVAE was trained on a single NVIDIA A100 (80GB) GPU, with each epoch requiring approximately 30 minutes. We utilized the AdamW optimizer with a base learning rate of $5 \times 10^{-4}$ for 100 epochs. To prioritize 3D backbone reconstruction, we used asymmetric loss weights: coordinates (10.0), angles (5.0), sequence (1.0), and KL (0.1). As illustrated in Figure~\ref{fig:training_dynamics} (a), the total validation loss decays rapidly during the initial phase and reaches a stable plateau around 80 epochs.

\textbf{Stage 2: Latent Diffusion Training.}
Following the latent manifold, the diffusion network was trained on the latent space. Under the same hardware environment, each epoch of the Latent Diffusion stage required approximately 1 hour. The model was trained for 100 epochs with a learning rate of $1 \times 10^{-4}$ and a batch size of 32. As shown in Figure~\ref{fig:training_dynamics} (b), the overall training loss exhibits a smooth decay, confirming the stability of the diffusion process within the compressed latent space.

\textbf{Computational Cost and Inference Latency.}
 Using the latent compression strategy, GOLD avoids the computational explosion commonly found in all-atom diffusion models. The entire pipeline achieves convergence within roughly 100 A100-hours. During inference, generating a single RNA 3D structure takes only a few seconds (please see details in appendix ~\ref{app:inference_efficiency}), demonstrating potential for high-throughput RNA design tasks.

 \section{Experimental Details}
\label{app:experimental_setup}

\subsection{Dataset Construction}
\label{app:dataset_construction}

\textbf{GOLD Benchmark:} To train the GOLD and conduct inverse folding evaluations, we constructed the high-quality GOLD Benchmark. Unlike studies that simply use raw PDB files directly, we designed an optimized workflow to obtain high-quality data. The detailed procedures are as follows:
\begin{enumerate}
    \item \textbf{Initial Collection and Filtering:} We aggregated all publicly available 3D structures containing RNA from the Protein Data Bank (PDB) released up to August 19, 2025. Initially, we excluded short fragments with a sequence length of less than 5 nt, as well as low-quality structures with a resolution worse than 4.0 \text{\AA}.
    \item \textbf{RNA Entity Definition and Topology Extraction:} For downstream processing, we defined fundamental RNA entities by clustering RNA chains according to inter-chain base-pairing interactions identified by FR3D ~\cite{sarver2008fr3d}. Specifically, we generated an RNA-chain interaction graph where an edge between two chains was established if they shared at least three inter-chain base pairs, and this inter-chain base-pair count amounted to no less than $15\%$ of the shorter chain's length. Groups of connected RNA chains in this graph were merged into single integrated RNA entities, whereas isolated chains lacking such links were treated as standalone entities. The ground-truth secondary structures (gt ss) were derived simultaneously from this FR3D pairing network. These RNA entities served as the fundamental units for all subsequent preprocessing procedures.
     \item \textbf{Multi-modal Consistency Check:} Since raw PDB files frequently contain missing residues or mismatched annotations, we re-parsed the crystalline coordinates to extract physically present nucleotides (\texttt{seq}), which were aligned against the ground-truth reference sequence. Furthermore, for the RNA entities defined above, we enforced consistency validation between the 1D sequence, 2D topology and 3D coordinates. Any sample exhibiting mismatches in length or composition in these modalities was classified as noise and filtered out.
    \item \textbf{Chemical Modification Normalization:} Native nucleic acids are rich in complex chemical modifications. To align with the standard 4-letter discrete vocabulary (A, U, C, G) required by foundation models and our GeoVAE, we implemented a precise mapping dictionary to revert non-standard modified residues to their canonical counterparts. 
     \item \textbf{Dual-level Homology Isolation and Splitting:} To definitively prevent data leakage caused by homologous sequences across the evaluation sets, we applied rigorous clustering across two independent representation spaces. This prevents highly similar samples from being distributed across different data splits (train, validation, or test):
    \begin{itemize}
        \item \textbf{Sequence-level:} We utilized MMseqs2 for sequence clustering, setting the sequence identity threshold to a stringent 0.3 ($30\%$).
        \item \textbf{Structural-level:} We further utilized US-align to calculate 3D topological alignment scores, clustering structures with a similarity (TM-score) exceeding 0.45.
    \end{itemize}
    Any samples with the sequence identity ($>30\%$) or structural similarity (TM-score $>0.45$) exceeding thresholds were assigned to the same cluster. During dataset partitioning, each cluster was treated as an indivisible unit and assigned entirely to a specific split. This strategy eliminates homologous information leakage between the training and evaluation sets, ensuring that the evaluation of generative models is reliable.
\end{enumerate}
The non-redundant core dataset selected through this rigorous pipeline was split into training, validation, and testing sets (7583/1144/598). Importantly, to ensure fair benchmarking, the extracted ground-truth secondary structure (ss) information was strictly excluded from the training phase of all evaluated algorithms. Please see details in Appendix ~\ref{app:ss_leakage_ablation}. This topology information was exclusively reserved for downstream evaluation metrics and zero-shot inference guidance.

\textbf{CASP15 Targets:} To evaluate the out-of-distribution (OOD) generalization capability on novel structures, we further compiled an independent test set from the CASP15 challenge. This challenging set comprises 60 sequence-structure pairs, providing a benchmark for assessing the model's robustness on unseen landscapes.

\subsection{Baseline Implementations}
\label{app:baseline_implementations}

To ensure a fair comparison, all baseline models were trained and evaluated on our standardized GOLD Benchmark dataset splits. We adapted the models originally designed for proteins to accommodate RNA structures. The specific implementation details are as follows:

\textbf{PiFold:} 
PiFold was originally proposed as a highly efficient graph neural network architecture for protein structure design \cite{gao2023pifold}. To fairly evaluate this model on RNA inverse folding task, we performed specific domain adaptation based on its official open-source implementation (\url{https://github.com/A4Bio/PiFold}) to accommodate nucleic acids. The specific modifications are as follows:
\begin{itemize}
    \item \textbf{Atomic Representation Substitution:} We replaced the amino acid backbone atoms (N, CA, C, O) relied upon for spatial $k$-NN graph construction with key atoms representing the coarse-grained RNA backbone (namely, the C1' atom of the phosphate backbone, the C4' atom of the ribose ring, and the N1/N9 pseudo-atoms representing the spatial orientation of the nucleobases).
    \item \textbf{Feature:} At the feature engineering level, we retained PiFold's core "Virtual Center" node representation mechanism, as well as the computational logic for multi-spatial edge features, which comprise multi-dimensional pairwise distances, dihedral angles, and spatial orientations.
    \item \textbf{Network and Output Adaptation:} The model backbone remains composed of multiple layers of \texttt{ProDesign\_Encoder} and \texttt{ProDesign\_Decoder}, aggregating local geometric information through graph message passing mechanism. At the output layer, we modified the classification head to predict the probability distribution over the four standard RNA nucleotides (A, U, C, G) instead of the 20 canonical amino acids.
\end{itemize}

During the training phase, we utilized the exact same training and validation splits as our GOLD benchmark to ensure the comparability of the results, and optimized the parameters end-to-end using the standard CrossEntropyLoss.

\textbf{RDesign:} 
RDesign is a native graph neural network architecture explicitly developed for RNA 3D structure inverse folding ~\cite{tan2024rdesign}. To ensure a fair evaluation on our GOLD Benchmark, we seamlessly integrated its official open-source implementation (\url{https://github.com/A4Bio/RDesign}) into our standardized training and evaluation pipeline. 

RDesign constructs spatial $k$-NN graphs using coarse-grained atoms. We configured the model with a hidden dimension of 128, along with 3 encoder and 3 decoder layers. It predicts the categorical probability distribution over the four nucleotides. For a rigorous comparison, RDesign was trained on our GOLD dataset splits using the AdamW optimizer (with a base learning rate of $3 \times 10^{-4}$) with a batch size of 8 for 100 epochs, minimizing the standard CrossEntropyLoss.

\textbf{gRNAde:} 
gRNAde is an autoregressive RNA inverse folding model utilizing Geometric Vector Perceptrons (GVPs) \cite{joshi2025grnade}. Based on its open-source implementation (\url{https://github.com/chaitjo/geometric-rna-design}), we constructed spatial graphs using coarse-grained RNA atomic coordinates (capped at a 4.5 \AA \ radius and 32 nearest neighbors), extracting multi-dimensional scalar and vector features (including 32-bin RBF and 32-dim positional encodings). We configured the model with 4 GVP convolutional layers, node hidden dimensions of 128 scalar and 16 vector, edge hidden dimensions of 64 scalar and 4 vector. For a rigorous comparison, gRNAde was trained on our GOLD dataset splits. We optimized the model using the Adam optimizer (with a base learning rate of $1 \times 10^{-4}$) and a label smoothing factor of 0.05, employing dynamic batch sampling (maximum 3000 nodes per batch) for 100 epochs.

\textbf{RiboDiffusion:} 
RiboDiffusion is a continuous-time diffusion model designed for RNA inverse folding \cite{han2024ribodiffusion}. Since its official open-source repository (\url{https://github.com/ml4bio/RiboDiffusion}) lacks training scripts, we implemented a custom training pipeline from scratch to enable training on our GOLD dataset splits. Architecturally, the model constructs spatial $k$-NN graphs ($k=10$) and extracts 3D geometric features via 4 Geometric Vector Perceptron (GVP) convolutional layers (node hidden dimensions configured to 512 scalar and 128 vector). These representations are subsequently fed into an 8-layer Transformer equipped with 16 attention heads for sequence denoising. We utilized a cosine noise schedule and optimized the network using the AdamW optimizer (with a base learning rate of $1 \times 10^{-4}$ and a weight decay of 0.001) and a batch size of 1 with 8 gradient accumulation steps over 1000 training epochs.

\textbf{SOLD:} 
SOLD is a latent diffusion network that leverages sequence foundation models for inverse folding \cite{si2026structure}. Based on its official open-source repository (\url{https://github.com/darkflash03/SOLD}), we reproduced its two-stage training pipeline on our GOLD dataset splits. In the first stage (feature compression), an MLP autoencoder compresses 640-dimensional RNA-FM pre-trained sequence embeddings into a 32-dimensional latent space, paired with a 3-layer MLP decoder (dropout 0.1) to predict nucleotide probabilities. This stage was trained for 100 epochs using the Adam optimizer (base learning rate $1 \times 10^{-4}$) with a batch size of 1 and 16 gradient accumulation steps. In the second stage (latent diffusion), the model executes a continuous-time diffusion process over 100 denoising steps. The denoising architecture combines 4 GVP convolutional layers (node hidden dimensions configured to 512 scalar and 128 vector) and an 8-layer Transformer equipped with 16 attention heads. This diffusion stage was trained end-to-end for 100 epochs using the AdamW optimizer (learning rate $1 \times 10^{-4}$, weight decay 0.01) with a batch size of 64.

\subsection{Evaluation Metrics}
\label{app:evaluation_metrics}

To comprehensively evaluate the performance of  generative models in sequence, topology, and spatial dimensions, we formalized the mathematical definitions of all evaluation metrics in our pipeline. Notably, for secondary structure evaluation, we engineered two independent computational protocols tailored to different experimental objectives.

\textbf{Sequence Recovery (Seq Recovery):}
Measures the proportion of exact character matches between the generated sequence $\hat{S}$ and the ground truth sequence $S^{gt}$ at each aligned position over length $L$:
\begin{equation}
    \text{Seq Recovery} = \frac{1}{L} \sum_{i=1}^L \mathbb{I}(\hat{s}_i = s^{gt}_i)
\end{equation}

\textbf{Secondary Structure Recovery (SS Recovery):}
In the \textbf{Main Results} section, to fairly evaluate the intrinsic topological potential of the generated sequences while eliminating the inherent prediction bias of external tools, we input both the generated sequence $\hat{S}$ and the ground truth sequence $S^{gt}$ into the ViennaRNA folding engine. We then compute the character-level match between their full-length dot-bracket strings:
\begin{equation}
    \text{SS Recovery} = \frac{1}{L} \sum_{i=1}^L \mathbb{I}\left( \text{ViennaRNA}(\hat{S})_i = \text{ViennaRNA}(S_{gt})_i \right)
\end{equation}

\textbf{RMSD based on RhoFold \& AlphaFold3:}
To evaluate the 3D physical foldability of the generated sequences, we employed RhoFold and AlphaFold3 (AF3) as folding oracles. For RhoFold, we executed the official pretrained weights in \texttt{single\_seq\_pred} mode locally. For RNA structure prediction using AF3, five predictions were generated using a fixed random seed (seed = 2026), and only the top-ranked poses selected by ranking score. All other parameters remained at their default values.

Given that sequence-based alignment fails in inverse folding due to sequence mutations, we developed a global RMSD algorithm based on the backbone C4' atoms that considers only the topological structure. Given the C4' coordinates of the ground truth structure $X \in \mathbb{R}^{L \times 3}$ and the predicted structure $Y \in \mathbb{R}^{L \times 3}$, we map the residues by position and align them to the origin ($\bar{X} = X - \mu_X, \bar{Y} = Y - \mu_Y$). The covariance matrix $H = \bar{Y}^T \bar{X}$ undergoes Singular Value Decomposition (SVD) to analytically solve for the optimal Kabsch rotation matrix ~\cite{kabsch1976solution} $R$:
\begin{equation}
    R = V \begin{pmatrix} 1 & 0 & 0 \\ 0 & 1 & 0 \\ 0 & 0 & \det(VU^T) \end{pmatrix} U^T
\end{equation}
The RMSD is then calculated:
\begin{equation}
    \text{RMSD} = \sqrt{ \frac{1}{L} \sum_{i=1}^L \left\| \bar{X}_i - (\bar{Y}_i R) \right\|^2 }
\end{equation}

\textbf{SS Validity in Ablation Analysis:} 
Unlike our main experiments that rely on external secondary structure predictors, analyzing the ablation of the topological guidance parameter ($\omega_{ss}$) requires a tool-agnostic perspective to isolate its direct impact. To avoid any biases introduced by external prediction tools, we propose a deterministic physical verification metric. Based on the ground-truth base-pairing graph $E_{ss}$ extracted directly from the native 3D structure, this metric strictly verifies whether the generated sequence successfully satisfies Watson-Crick (A-U, G-C) or Wobble (G-U) base-pairing rules at the designated topological edges:
\begin{equation}
    \text{SS Validity} = \frac{1}{|E_{ss}|} \sum_{(i,j) \in E_{ss}} \mathbb{I}\left((\hat{s}_i, \hat{s}_j) \in \mathcal{P}_{valid}\right)
\end{equation}
where $\mathcal{P}_{valid} = \{(\text{A,U}), (\text{U,A}), (\text{G,C}), (\text{C,G}), (\text{G,U}), (\text{U,G})\}$. This direct metric provides an clear perspective for evaluating the effectiveness of topological guidance.

\textbf{Clash Score in Ablation Analysis:}
Standard structural evaluation pipelines often employ coordinate relaxation tools, which automatically resolve atomic clashes post-hoc. Relying on such post-processing would inevitably obscure the direct contribution of our spatial energy guidance. To independently analyze the inherent capabilities of our guidance mechanism in ablation studies, we directly calculate the Clash Score on the unrelaxed 3D coordinates generated by our model. This metric is defined as the number of non-bonded atomic pairs with a spatial distance below $2.0\,\text{\AA}$ per 1000 atoms. To avoid falsely penalizing natural covalent bonds, we strictly mask the distances between adjacent residues (i.e., $|i - j| \le 1$):
\begin{equation}
    \text{Clash Score} = \frac{1000}{N_{atoms}} \sum_{i, j: |i - j| > 1} \mathbb{I}(d(i,j) < 2.0\,\text{\AA})
\end{equation}

\section{Performance and Cost Analysis on Sequence Lengths}
\label{app:long_sequence}

In RNA 3D inverse folding, the solution space for geometric modeling and sequence generation expands rapidly as sequence length increases. To evaluate the model's performance and computational scalability on extended sequences, we conducted an analysis from two aspects: generation quality and inference cost.

\subsection{Performance across RNA of different lengths.}
\label{app:long_rna}
To compare generation quality, we partitioned the GOLD Benchmark test set into short-to-medium sequences ($L \le 100$) and long sequences ($L > 100$). As shown in Table~\ref{tab:long_sequence_perf}, the transition to the $L > 100$ regime poses an extreme challenge for the entire field, evidenced by a significant degradation in RMSD across all baseline models. However, under these challenging conditions, GOLD exhibits the strongest relative robustness. While maintaining leading structure metrics, GOLD successfully achieves high Sequence Recovery and outperforms the strongest baseline (SOLD) in 2D topological pairing accuracy. 

\begin{table}[ht]
    \centering
    \caption{Performance breakdown across different sequence length regimes on the GOLD Benchmark.}
    \label{tab:long_sequence_perf}
    \small 
    \setlength{\tabcolsep}{4.5pt} 
    \begin{tabular}{l cccc}
        \toprule
        \multirow{2}{*}{\textbf{Method}} & \multirow{2}{*}{\textbf{Seq Recovery ($\uparrow$)}} & \multirow{2}{*}{\textbf{SS Recovery ($\uparrow$)}} & \multicolumn{2}{c}{\textbf{3D RMSD (\AA, $\downarrow$)}} \\
        \cmidrule(lr){4-5}
        & & & \textbf{RhoFold} & \textbf{AlphaFold3} \\
        \midrule
        
        \multicolumn{5}{c}{\cellcolor{gray!15}\textbf{Short-to-Medium Sequences ($L \le 100$)}} \\
        \midrule
        PiFold         & $0.4237$ & $0.6289$ & $15.81$ & $17.36$ \\
        RDesign        & $0.4443$ & $0.6918$ & $13.06$ & $13.47$ \\
        gRNAde          & $0.5440$ & $0.6614$ & $13.75$ & $14.38$ \\
        RiboDiffusion  & $0.5158$ & $0.7061$ & $11.56$ & $12.41$ \\
        SOLD           & $0.5837$ & $0.7479$ & $11.07$ & $12.37$ \\
        \midrule
        \textbf{GOLD (Dual-Guided)} & $\mathbf{0.6397}$ & $\mathbf{0.8332}$ & $\mathbf{8.968}$ & $\mathbf{10.80}$ \\
        
        \midrule
        \midrule
        
        \multicolumn{5}{c}{\cellcolor{gray!15}\textbf{Long Sequences ($L > 100$)}} \\
        \midrule
        PiFold         & $0.3776$ & $0.4877$ & $38.37$ & $37.19$ \\
        RDesign        & $0.4245$ & $0.4554$ & $36.15$ & $37.39$ \\
        gRNAde         & $0.5609$ & $0.4302$ & $37.46$ & $39.33$ \\
        RiboDiffusion  & $0.4684$ & $0.5445$ & $34.41$ & $32.95$ \\
        SOLD           & $0.5839$ & $0.5868$ & $34.49$ & $33.01$ \\
        \midrule
        \textbf{GOLD (Dual-Guided)} & $\mathbf{0.6289}$ & $\mathbf{0.6237}$ & $\mathbf{33.43}$ & $\mathbf{32.67}$ \\
        \bottomrule
    \end{tabular}
\end{table}

\subsection{Inference Efficiency Across Different Lengths.}
\label{app:inference_efficiency}

To evaluate the scalability of our approach in practical scenarios, we measure the generation latency and peak GPU memory consumption across four RNA targets with lengths ($L$) ranging from $44$ to $308$. All tests are conducted in a unified hardware environment using a single NVIDIA A100 (80GB) GPU. 

As detailed in Table~\ref{tab:long_sequence_cost}, compressing the generation process into a low-dimensional latent manifold effectively avoids the heavy memory bottlenecks typically associated with explicit spatial coordinate generation. Consequently, \textbf{GOLD (Base)} demonstrates highly competitive computational efficiency: it achieves comparable inference cost to the unimodal baseline SOLD, while maintaining a peak GPU of only around $2$ GB even for sequences longer than 300 nucleotides. 

Furthermore, applying the dual-guided mechanism introduces negligible GPU overhead across all lengths. In terms of inference speed, calculating the physical constraint gradients at each denoising step incurs only a minimal time penalty for shorter RNA sequences. Although this inference cost inevitably scales for longer targets, the absolute time remains well within practical limits for downstream RNA design tasks. In summary, these results demonstrate that GOLD provides a scalable architecture that effectively balances structural generation quality with computational efficiency.

\begin{table}[ht]
    \centering
    \caption{Inference cost across RNA targets. Time (T) in seconds, Peak Memory (M) in MB.}
    \label{tab:long_sequence_cost}
    \footnotesize 
    \setlength{\tabcolsep}{5pt}
    \begin{tabular}{lr ccc ccc}
        \toprule
        \multirow{2}{*}{\textbf{PDB ID}} & \multirow{2}{*}{\textbf{Len}} & \multicolumn{3}{c}{\textbf{Inference Time (s) $\downarrow$}} & \multicolumn{3}{c}{\textbf{Peak GPU Mem (MB) $\downarrow$}} \\
        \cmidrule(lr){3-5} \cmidrule(lr){6-8}
        & & \textbf{SOLD} & \textbf{GOLD$_{Base}$} & \textbf{GOLD$_{Dual}$} & \textbf{SOLD} & \textbf{GOLD$_{Base}$} & \textbf{GOLD$_{Dual}$} \\
        \midrule
        1A60\_A & 44   & 5.12 & 5.37 & 7.36  & 1486 & 1525 & 1529 \\
        8QO4\_A & 156  & 5.34 & 5.93 & 9.69  & 1598 & 1693 & 1708 \\
        5TC1\_R & 242  & 6.28 & 6.85 & 11.61 & 1786 & 1811 & 1847 \\
        6N7X\_R & 308  & 7.05 & 7.61 & 14.78 & 1876 & 2043 & 2088 \\
        \bottomrule
    \end{tabular}
\end{table}

\section{Why SO(2) Layer is Necessary in GeoVAE ?}
\label{app:so2_rationale}

In Section ~\ref{GeoVAE} of the main text, we introduced a $SO(2)$-based rotation-aware architecture for the GeoVAE encoder and decoder. In this section, we evaluate the advantages of this design through comparisons with alternative sequence--torsion fusion strategies. We first analyze feature-level sensitivity and training dynamics against a traditional concatenation baseline, followed by a quantitative comparison with a stronger Fourier--FiLM baseline.

\subsection{Experimental Setup}
\label{app:so2_poc}

To isolate the effects of the fusion mechanism, we trained two models under identical settings for the perturbation and training-dynamics analyses: our \textbf{$SO(2)$ GeoVAE} and a baseline model that directly concatenates sequence embeddings and trigonometric angle vectors (\textbf{Trig-Concat Baseline}). Both models were supervised exclusively by sequence and angle reconstruction losses. The subsequent quantitative fusion comparison adopts the complete GeoVAE training objective, including coordinate reconstruction supervision.

\textbf{Evaluation Metric}. To evaluate the structural reconstruction accuracy, we employ the Kinematic Approximation Error (\textbf{KAE}) in Section ~\ref{preliminary}. This metric provides a direct geometric assessment of the decoded features.

\subsection{Mitigating Modality Collapse}
\label{app:so2_observation_1}
To probe how each architecture processes multi-modal features internally, we conducted an input-level spatial perturbation test. We selected a target RNA nucleotide from the GOLD Benchmark test set, systematically perturbed its input torsion angle from $-\pi$ to $\pi$, and recorded the corresponding changes in the predicted sequence probability ($\hat{P}_{seq}$).

\begin{figure}[htbp]
    \centering
    \includegraphics[width=0.85\textwidth]{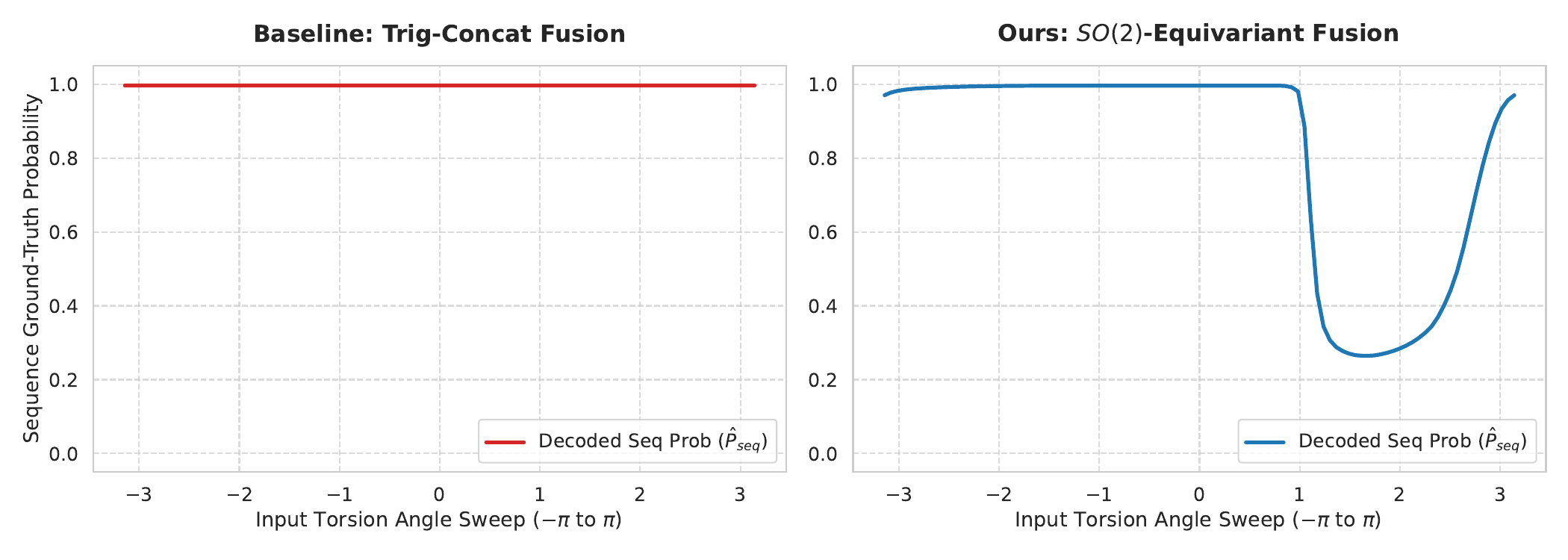} 
    \caption{\textbf{modality collapse vs. synergistic coupling.} Comparison of predicted sequence probability ($\hat{P}_{seq}$) in response to a continuous input-level angular sweep. The Trig-Concat Baseline (left) exhibits an entirely flat response, indicating that the network zeroed out spatial weights early in training. In contrast, the $SO(2)$ GeoVAE (right) demonstrates sensitivity to geometric perturbations, proving that its architecture prevents information truncation and forces the synergistic coupling of sequence and geometry.}
    \label{fig:spatial_perturbation}
\end{figure}

As illustrated in Figure~\ref{fig:spatial_perturbation} (left), the sequence probability predicted by the Trig-Concat baseline remains largely invariant regardless of how the input angle is manipulated. This observation suggests that the Trig-Concat model exhibits a form of modal collapse, likely due to an optimization imbalance between discrete sequence classification and continuous angle regression. To rapidly minimize the total loss early in training, the network relies primarily on the discrete sequence features, thereby ignoring the continuous spatial inputs.

Conversely, the $SO(2)$ model demonstrates consistent responsiveness to spatial perturbations (Figure~\ref{fig:spatial_perturbation}, right). In the $SO(2)$ architecture, sequence (magnitude) and angles (phase) are explicitly coupled through complex multiplication. Altering the input angle inherently shifts the spatial context processed by the subsequent Transformer layers, predictably modifying the sequence output. This structural coupling prevents the network from bypassing geometric cues, thereby encouraging a more reliable integration of sequence and spatial constraints.

\subsection{Structural Optimization Dynamics}
\label{app:so2_observation_2}

\begin{figure}[htbp]
    \centering
    \includegraphics[width=0.65\textwidth]{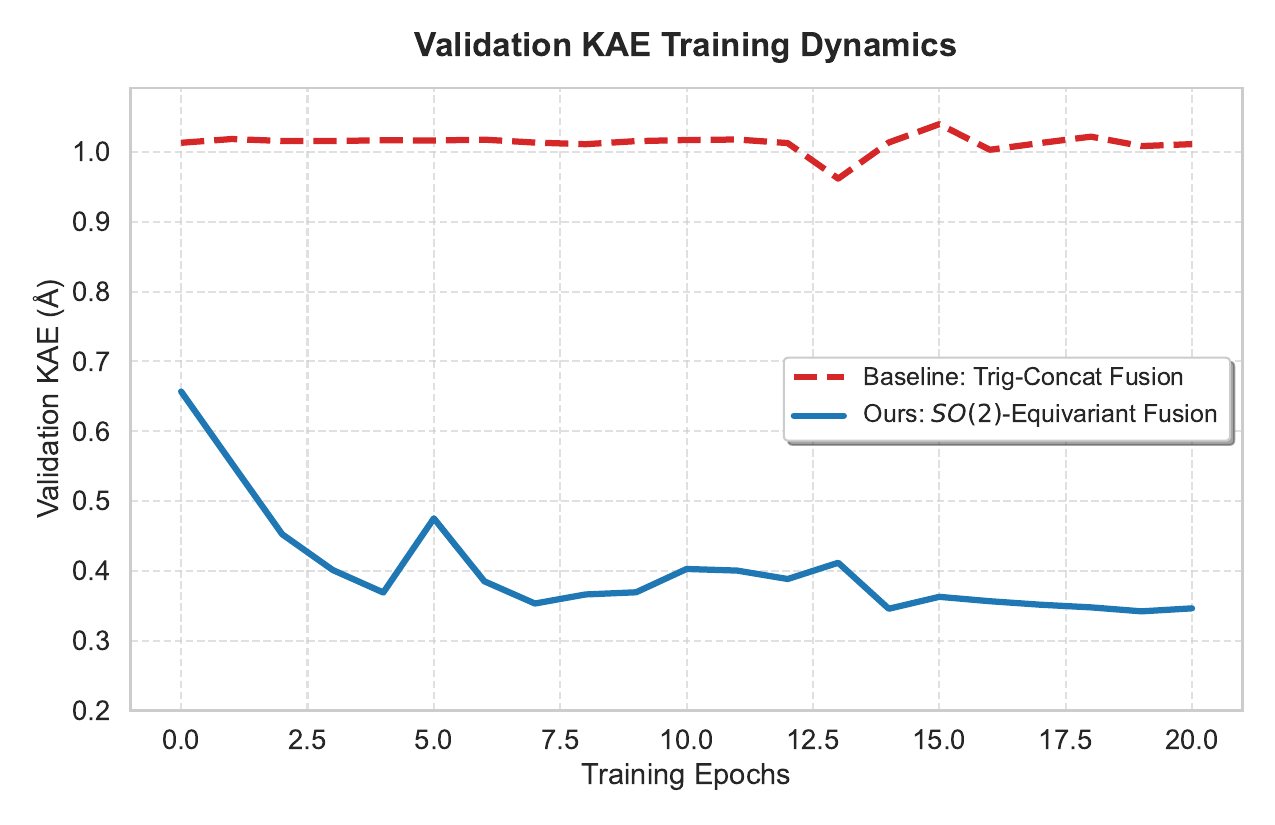} 
    \caption{\textbf{Structural optimization efficiency.} Comparison of the validation KAE descent between the two fusion strategies. The Trig-Concat baseline gets trapped in an optimization plateau near $1.0$~\AA\ because it ignores spatial features, whereas the $SO(2)$ model demonstrates a smooth and stable structural convergence.}
    \label{fig:training_dynamics_so2}
\end{figure}

These differences in feature processing directly affect the overall training dynamics of the two models. As shown in Figure ~\ref{fig:training_dynamics_so2}, there are significant differences in their structural optimization approaches. Because the Trig-Concat baseline relies heavily on discrete sequence features and largely ignores continuous spatial inputs during the early stages of training (as shown by the perturbation test), its initial structural predictions lack geometric guidance. Consequently, its KAE gets stuck in an optimization plateau near $1.0$ \AA\ for over 20 epochs. 

In contrast, the $SO(2)$ GeoVAE avoids this issue through its effective coupling. By forcing the network to process both 3D geometry and sequence semantics simultaneously throughout the training, the KAE of the $SO(2)$ model decreases smoothly, leading to a stable and efficient structural convergence.

\subsection{Quantitative Comparison of Sequence--Torsion Fusion Strategies}
\label{app:so2_fusion_ablation}

To further validate the effectiveness of the $SO(2)$-aware fusion mechanism, we extend the comparison beyond Trig-Concat by introducing a stronger periodic fusion baseline, \textbf{Fourier--FiLM}~\cite{perez2018film}. Specifically, each torsion angle $\chi_i$ is encoded using $K=4$ Fourier frequencies, $[\sin(k\chi_i),\cos(k\chi_i)]_{k=1}^{K}$, and mapped to feature-wise scale and shift vectors $\boldsymbol{\gamma}_i,\boldsymbol{\beta}_i\in\mathbb{R}^{256}$. These vectors modulate the sequence features through
\begin{equation}
    \widetilde{\mathbf{h}}_i =
    (1+\boldsymbol{\gamma}_i)\odot\mathbf{h}_i
    +\boldsymbol{\beta}_i.
\end{equation}
We replace the $SO(2)$-aware fusion module in both the encoder and decoder while keeping the remaining GeoVAE architecture, geometric losses, and training settings unchanged.

\begin{table}[ht]
    \centering
    \caption{Comparison of sequence--torsion fusion strategies in GeoVAE on the GOLD Benchmark.}
    \label{tab:so2_fusion_ablation}
    \small
    \setlength{\tabcolsep}{6pt}
    \begin{tabular}{lccc}
        \toprule
        \textbf{Fusion Strategy}
        & \textbf{Angle MSE ($\downarrow$)}
        & \textbf{Coord MSE ($\downarrow$)}
        & \textbf{KAE (\AA, $\downarrow$)} \\
        \midrule
        Trig-Concat
        & $0.1416$ & $0.3474$ & $1.0110$ \\
        Fourier--FiLM
        & $0.0173$ & $0.0517$ & $0.3659$ \\
        \textbf{$SO(2)$-aware}
        & $\mathbf{0.0109}$
        & $\mathbf{0.0287}$
        & $\mathbf{0.2466}$ \\
        \bottomrule
    \end{tabular}
\end{table}

As shown in Table~\ref{tab:so2_fusion_ablation}, Fourier--FiLM substantially improves geometric reconstruction over Trig-Concat, indicating the importance of periodic angle encoding and adaptive feature modulation. Nevertheless, the $SO(2)$-aware fusion consistently achieves the lowest angle MSE, coordinate MSE, and KAE. These results suggest that explicitly coupling sequence features and torsion angles through an $SO(2)$-structured representation provides a more effective inductive bias than conventional periodic feature fusion. Together with the perturbation and training-dynamics analyses above, this comparison further supports the effectiveness of the proposed $SO(2)$-aware architecture for joint sequence--geometry modeling.

\section{Discussion on Latent Dimension}
\label{app:latent_dim_tradeoff}

In image-based Latent Diffusion Models (LDMs), there is a widely discussed trade-off: higher latent dimensionality ($d$) yields better VAE reconstruction fidelity but often increases the difficulty of training the downstream diffusion model ~\citep{ramanujan2025when}. However, in our RNA inverse folding task, we observed empirical phenomena that somewhat differ from the findings of conventional image-domain findings. This section discusses the interplay between GeoVAE information compression, DiT generative bandwidth, and training dynamics.

\subsection{Decoupling Reconstruction Fidelity from Generative Capacity}
\label{app:reconstruction_vs_generation}

\begin{table}[htbp]
    \centering
    \caption{\textbf{Evaluation of latent dimension ($d$) scaling.} We compare the GeoVAE's upper-bound reconstruction capacity with the actual generative performance of Latent Diffusion Model. To isolate the impact of latent dimensionality, all generative metrics are evaluated on the \textbf{GOLD Benchmark under the unguided setting}. GeoVAE achieves near-perfect local reconstruction even at $d=4$, indicating a low intrinsic physical dimension. However, Latent Diffusion Model generation peaks at $d=8$. Further scaling up to $d=64$ plateaus without causing severe performance degradation. Best generative results are highlighted in \textbf{bold}.}
    \label{tab:latent_dim}
    \resizebox{\textwidth}{!}{
    \begin{tabular}{l cc p{0.5cm} ccc}
        \toprule
        & \multicolumn{2}{c}{\textbf{GeoVAE (Reconstruction)}} & & \multicolumn{3}{c}{\textbf{LDM (Generation)}} \\
        \cmidrule{2-3} \cmidrule{5-7}
        \textbf{Latent Dim ($d$)} & \textbf{Seq Recovery ($\uparrow$)} & \textbf{KAE ($\downarrow$)} & & \textbf{Seq Recovery ($\uparrow$)} & \textbf{SS Validity ($\uparrow$)} & \textbf{Clash Score ($\downarrow$)} \\
        \midrule
        $d = 4$  & 0.9998 & 0.2465 & & 0.5957 & 0.8747 & 5.41 \\
        \rowcolor{gray!10} 
        \textbf{$d = 8$} & 0.9996 & 0.2466 & & \textbf{0.6249} & \textbf{0.9040} & \textbf{4.51} \\
        $d = 16$ & 0.9997 & 0.2397 & & 0.6240 & 0.8955 & 4.54 \\
        $d = 32$ & \textbf{0.9999}& 0.2406 & & 0.6249 & 0.9033 & 4.58 \\
        $d = 64$ & 0.9998 & \textbf{0.2396}& & 0.6207 & 0.8924 & 4.63 \\
        \bottomrule
    \end{tabular}
    }
\end{table}

As shown in Table ~\ref{tab:latent_dim}, compressing the latent space to $d=4$ maintains near-perfect GeoVAE reconstruction accuracy (Sequence Recovery $\approx 1.0$, KAE $\approx 0.24$\AA). We attribute this to the extremely low \textit{intrinsic manifold dimension} of RNA local interactions. The physical degrees of freedom of a single nucleotide consist solely of a discrete identity state ($\textasciitilde2$ bits) and backbone torsion constrained by steric hindrance. A continuous 4-dimensional complex space provides a state capacity that exceeds this local information bottleneck. 

However, although $d=4$ satisfies the  capacity required for VAE, the generative performance of the LDM still improves at $d=8$. This phenomenon highlights the decoupling of local reconstruction capacity and the complexity of diffusion. In the VAE phase, forcing features into an aggressively compressed $d=4$ space results in a highly entangled and densely packed latent manifold. Although the VAE decoder can deterministically map these entangled representations back to structures, the diffusion model faces a harder optimization task.

During the iterative denoising process, the diffusion model must approximate the score vector field ($\nabla_z \log p_t(z_t)$) to construct a generative trajectory \cite{song2019generative}. In an over-compressed $d=4$ space, the latent manifold is steeply non-linear, making the score function highly sensitive to Gaussian noise perturbations~\cite{vahdat2021score}. Expanding the dimensionality to $d=8$ effectively unrolls the latent manifold into a slightly higher-dimensional space ~\cite{bengio2013representation}. This unrolling reduces feature entanglement and provides a smoother score field for the diffusion model to approximate, ultimately stabilizing the generative spatial routing without incurring the curse of dimensionality.

\subsection{Robustness to Latent Dimension Scaling}
\label{app:explain_latent_dim_robustness}
To explore the limits of latent dimensional scaling, we evaluated hyper-dimensional latent up to $d=64$. In visual domains, over-parameterized latent typically encode unstructured high-frequency artifacts, leading to KL divergence explosion and catastrophic generative degradation~\citep{ramanujan2025when}. However, in RNA inverse folding task, our model exhibits a robust performance plateau without significant degradation in generative performance  (Table~\ref{tab:latent_dim}).

In standard continuous generation tasks (e.g., image synthesis), scaling up the latent dimension typically exacerbates the rate-distortion tradeoff ~\citep{ramanujan2025when} in VAEs. To achieve better reconstruction results by utilizing higher dimensions, the model must encode more information, which significantly increases the KL divergence. This creates a severe mismatch with the standard Gaussian prior, which ultimately affects the quaility of downstream generative sampling. Conversely, if the KL penalty is strictly enforced, the model often suffers from severe posterior collapse, rendering the extra dimensions useless. 

\begin{figure}[htbp]
    \centering
    \includegraphics[width=1.0\textwidth]{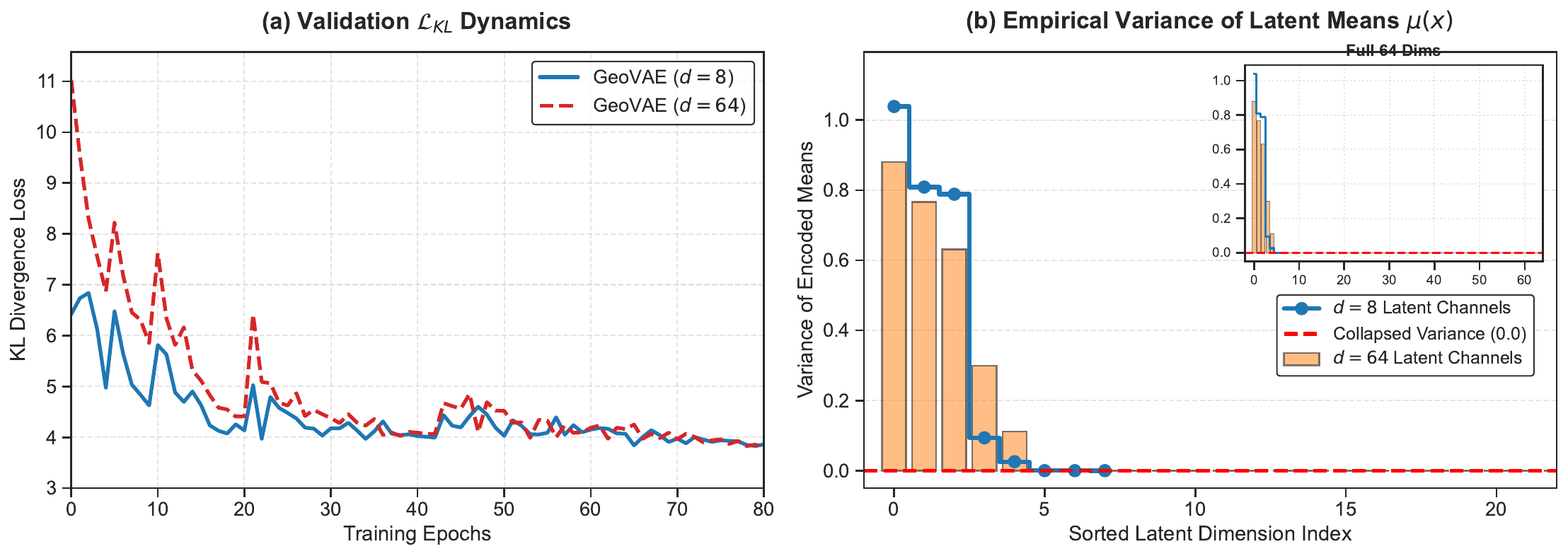} 
    \caption{\textbf{Comparative evidence of Partial Posterior Collapse.} (a) The KL divergence loss remains stable across 80 epochs even when channel capacity is scaled $d=8$ to 64. (b) Comparative analysis of the sorted empirical variance of latent means ($\mu(x)$). The \textbf{main plot} provides a zoomed-in view of the active dimensions, revealing that the variance profile of the over-parameterized $d=64$ model perfectly mirrors the intrinsic active dimensions of the baseline $d=8$ model. This confirms that all remaining redundant channels collapse to zero (red dashed line). A zero variance of encoded means indicates that the encoder outputs a constant regardless of the input, proving that these excess dimensions encode no mutual information and are entirely suppressed by the posterior collapse mechanism.}
    \label{fig:posterior_collapse}
\end{figure}

However, the training curves (Figure~\ref{fig:posterior_collapse}(a)) reveal a fundamentally different behavior in GOLD. Although channel capacity increased, the KL divergence loss ($\mathcal{L}_{KL}$) of the $d=64$ model does not inflate; rather, it converges stably to a magnitude comparable to the baseline $d=8$ setting while maintaining high-fidelity reconstruction. 

We attribute this harmony to the inherently low intrinsic dimensionality of RNA structures. Unlike image pixels, RNA conformations are strictly governed by rigid physical and topological rules. Because our $SO(2)$-equivariant manifold efficiently encapsulates these geometric dependencies within a low-dimensional subspace, the excess latent dimensions in the $d=64$ model are not forced to encode complex variances. Mathematically, this implies that the inactive dimensions contribute virtually zero to the cumulative KL penalty. Consequently, GOLD can safely scale its channel capacity to facilitate a smoother diffusion process without compromising the critical prior alignment required for generation.

To prove this assumption, we conducted a comparative statistical analysis in the empirical variance of the encoded latent means ($\mu(x)$) on the GOLD benchmark training set (Figure~\ref{fig:posterior_collapse}(b)). In variational inference, if the encoder captures meaningful mutual information, the predicted means $\mu(x)$ will vary on different input samples, yielding an empirical variance greater than zero. Conversely, posterior collapse occurs when the encoder outputs a constant mean ($\mu(x) \to 0$) independent of the input, causing the empirical variance to approach zero.

By independently sorting the variances of both the $d=8$ and $d=64$ models, a striking geometric pattern emerges: the variance profile of the active dimensions in the $d=64$ model almost perfectly mirrors the intrinsic profile of the $d=8$ model. The additional dimensions encode zero mutual information, with the empirical variance of their means collapsing to $0$ (red dashed line).

This finding is consistent with the phenomenon of \textbf{Partial Posterior Collapse} \citep{bowman2015generating, lucas2019dont} in variational inference. Governed by the ELBO objective trade-off, the KL penalty actively suppresses the use of redundant capacity unless strictly necessary for reconstruction. Because RNA forms a bounded, low-frequency physical system with limited intrinsic degrees of freedom, the GeoVAE only requires a small number of core dimensions to encapsulate the intrinsic 3D geometric manifold. The surplus channels are disabled, causing the encoder to output constant zero means for those channels. Consequently, the downstream LDM merely treats these collapsed dimensions as meaningless noise, and its linear projection layers seamlessly attenuate them. This analysis confirms the robustness of our framework to latent dimension.

\section{Extended Ablation Studies and Generative Evaluation}
\label{app:extended_ablations}

In this section, we provide additional analyses of GOLD, including the advantages of the unified latent manifold over raw-space co-diffusion, the effectiveness of gradient guidance compared to direct secondary-structure conditioning, and the one-to-many generative properties of RNA design. We further investigate the interaction and sensitivity of the two physical guidance objectives to better understand their effects on the sampling process.

\subsection{Unified Latent Manifold vs Raw-Space Co-Diffusion}
\label{app:raw_space_comparison}

Recently, several pioneering studies have explored the co-design of RNA sequences and structures by directly applying diffusion models to the raw data spaces \citep{ma2025riboflow, li2026a}. These methods typically combine discrete categorical diffusion for sequences and continuous Gaussian diffusion for structures, linking them primarily through multi-task loss functions. Although these models are promising, we cannot deploy them directly as baseline models for two main reasons. First, our task focuses on inverse folding, which requires generating sequences conditioned on a given 3D backbone, whereas these co-design methods target different generation scenarios. Second, their source codes are currently not publicly available.

\begin{figure}[htbp]
    \centering
    \includegraphics[width=1.0\textwidth]{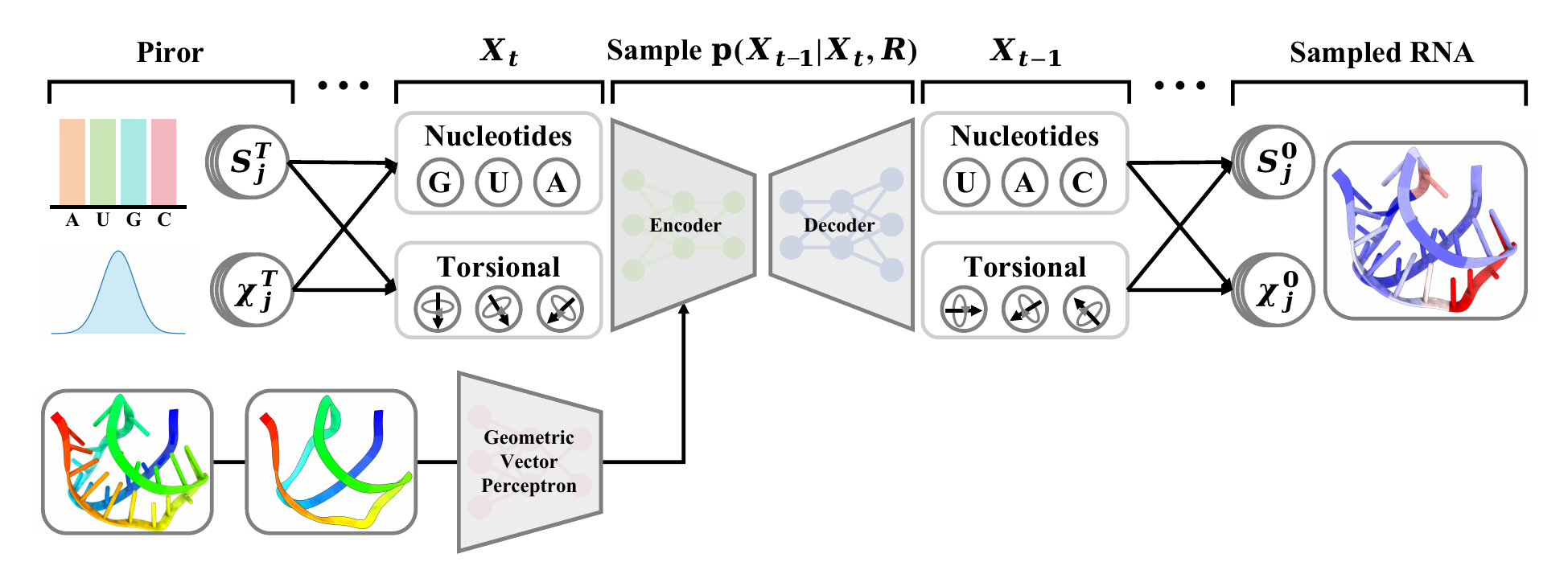} 
    \caption{\textbf{Architecture of the Raw-Space Co-Diffusion.} Unlike GOLD's unified latent manifold, this experiment performs independent diffusion processes (categorical and Gaussian) in their respective raw data spaces, fusing the noisy modalities solely through input-level concatenation.}
    \label{fig:raw_space_arch}
\end{figure}

\textbf{Raw-Space Co-Diffusion.} To analyze the structural impact of the latent manifold, we implemented a custom \textbf{Raw-Space Co-Diffusion Baseline}, as illustrated in Figure~\ref{fig:raw_space_arch}. To ensure a strictly controlled and fair comparison, this baseline retains the exact same generative backbone (GVP + DiT) as GOLD, but completely omits the GeoVAE. Instead of operating within a compressed manifold, it diffuses directly in the data space through the following mechanisms:

\begin{itemize}
    \item \textbf{Discrete Sequence Branch}: Applies categorical diffusion directly to the 4-letter nucleotide tokens, utilizing the exact same discrete transition kernels as our latent formulation.
    \item \textbf{Continuous Angle Branch}: Performs standard Gaussian diffusion on the trigonometric representations of the side-chain angles ($\chi$) within the raw coordinate space.
    \item \textbf{Input-side Fusion}: At each denoising timestep, the noisy sequence embeddings and noisy continuous angle vectors are simply concatenated at the input-side before being processed by the joint GVP-Transformer layers.
    \item \textbf{Joint Optimization}: The network is trained end-to-end to simultaneously minimize the raw-space data objectives, specifically cross entropy for sequence classification and MSE for angle regression.
\end{itemize}

\textbf{Experimental Results.} We evaluated both models on the internal GOLD Test Benchmark. To explicitly assess the native generation quality without relying on external folding oracles, we report the Sequence Recovery alongside two intrinsic structural metrics: SS Validity and Clash Score. As shown in Table~\ref{tab:raw_space_ablation}, while the Raw-Space Baseline achieves reasonable performance, its performance consistently worse than \textbf{GOLD (Base)} across all metrics. This performance gap indicates that simply combining parallel losses is insufficient for capturing the complex interdependencies of RNA. The unified latent manifold provides a smoother, physically entangled representation that effectively bridges the modality gap, natively suppressing steric clashes and improving topological pairing.

\begin{table}[ht]
    \centering
    \caption{Ablation study comparing the Unified Latent Manifold (GOLD) against the Raw-Space Co-Diffusion baseline on the GOLD Test Benchmark. All models are evaluated in the unguided (Base) setting using native outputs.}
    \label{tab:raw_space_ablation}
    \small
    \setlength{\tabcolsep}{8pt} 
    \begin{tabular}{l ccc}
        \toprule
        \textbf{Architecture} & \textbf{Seq Recovery ($\uparrow$)} & \textbf{SS Validity ($\uparrow$)} & \textbf{Clash Score ($\downarrow$)} \\
        \midrule
        Raw-Space Baseline & 0.6036 & 0.8657 & 6.53 \\
        \textbf{GOLD (Base)} & \textbf{0.6249} & \textbf{0.9040} & \textbf{4.51} \\
        \bottomrule
    \end{tabular}
\end{table}

\subsection{Input Feature vs Gradient Guidance}
\label{app:ss_leakage_ablation}

In our Dual-Guided framework, the generative trajectory is guided by the target 2D secondary structure (SS). To analyze the effectiveness of our gradient-based sampling mechanism from the availability of SS data, we conducted an ablation study to explore different methods of incorporating topological information.

\textbf{Ablation Setup: SS as an Input Feature.} We constructed a variant, \textbf{GOLD (Base w/ Input SS)}, which integrates the target SS graph directly into the initial spatial conditioning phase. Specifically, we explicitly added the target base-pairing edges to the spatial graph. In the edge attributes, edges connecting pairs of nucleotides were assigned a distinct binary indicator and this enriched graph was then processed by the GVP. In this setup, the model "sees" the exact same target topology as the guided version but relies entirely on implicit neural message passing to satisfy these constraints.

\textbf{Experimental Results and Analysis.} As shown in Table~\ref{tab:ss_leakage_ablation}, simply providing the target SS as input feature results in only a slight improvements compared to the GOLD (Base). This indicates that the network struggles to autonomously translate discrete input edge features into strict, long-range sequence pairing semantics during the latent denoising process. In contrast, \textbf{GOLD (SS-Guided)}, which enforces the exact same SS information through active gradient penalization ($\omega_{ss}=1.0$) at each sampling step, achieves a larger leap in both Sequence Recovery and SS Validity. This proves that the superior performance of our method stems from the explicit, dynamic mechanism of gradient guidance, rather than a trivial leakage of topological information.

\begin{table}[ht]
    \centering
    \caption{Ablation study on the utilization of Secondary Structure (SS) information evaluated on the GOLD Test Benchmark. Applying the target SS merely as an input edge feature yields minimal gains, demonstrating that the dynamic gradient intervention (SS-Guided) is the fundamental driver of topological accuracy.}
    \label{tab:ss_leakage_ablation}
    \small
    \setlength{\tabcolsep}{8pt}
    \begin{tabular}{l ccc}
        \toprule
        \textbf{Model Variant} & \textbf{Seq Recovery ($\uparrow$)} & \textbf{SS Validity ($\uparrow$)} & \textbf{Clash Score ($\downarrow$)} \\
        \midrule
        GOLD (Base) & 0.6249 & 0.9040 & 4.51 \\
        GOLD (Base w/ Input SS) & 0.6253 & 0.9045 & 4.48 \\
        \textbf{GOLD (SS-Guided)} & \textbf{0.6267} & \textbf{0.9343} & \textbf{4.32} \\
        \bottomrule
    \end{tabular}
\end{table}

\subsection{One-to-Many Generative Evaluation}
\label{app:one_to_many}

RNA inverse folding is inherently a one-to-many generation task, as multiple distinct sequences may be compatible with the same target structure. Therefore, conventional recovery metrics alone cannot fully characterize the generative capability of a design method. To provide a more comprehensive evaluation, we additionally assess sequence novelty, structural success rate as a proxy for designability, and intra-target sequence diversity, following evaluation practices in related studies~\citep{han2024ribodiffusion,joshi2025grnade,runge2024rnabench}.

\textbf{Evaluation Metrics.}
Sequence novelty measures the dissimilarity between generated sequences and the training set:
\begin{equation}
    \mathrm{Novelty}_{seq}
    =
    1-\frac{1}{n}\sum_{i=1}^{n}
    \max_{t\in\mathcal{T}}\mathrm{Id}(s_i,t),
\end{equation}
where $\mathcal{T}$ denotes the training set, $s_i$ is a generated sequence, and $\mathrm{Id}(s,t)$ is the sequence identity computed using Needleman--Wunsch global alignment.

Structural success rate measures the proportion of generated sequences whose predicted structures satisfy a given scRMSD threshold:
\begin{equation}
    \mathrm{SuccessRate}(\tau)
    =
    \frac{1}{n}\sum_{i=1}^{n}
    \mathbf{1}[\mathrm{scRMSD}_i\leq\tau],
\end{equation}
where $\tau\in\{2,5,8,10\}$\,\AA. While $2$\,\AA{} provides a strict structural criterion, we also consider broader thresholds to account for RNA flexibility and the limitations of current RNA structure prediction tools.

To evaluate sequence diversity, we compute the internal diversity (IntDiv) among the eight sequences generated for each target:
\begin{equation}
    \mathrm{IntDiv}(G_X)
    =
    1-\frac{1}{m_X^2}
    \sum_{i,j}\mathrm{Sim}(s_i,s_j),
\end{equation}
where $G_X$ is the set of $m_X=8$ generated sequences for target $X$, and $\mathrm{Sim}(s_i,s_j)=1-d_H(s_i,s_j)/L$ denotes normalized Hamming similarity. We report the average IntDiv across all test targets.

\begin{table}[ht]
    \centering
    \caption{Sequence novelty, structural success rate, and intra-target diversity on the GOLD Benchmark.}
    \label{tab:designability_diversity}
    \small
    \setlength{\tabcolsep}{5pt}
    \begin{tabular}{lcccccc}
        \toprule
        \textbf{Method}
        & \textbf{Novelty}
        & \multicolumn{4}{c}{\textbf{Success Rate ($\uparrow$)}}
        & \textbf{Diversity} \\
        \cmidrule(lr){3-6}
        & & \textbf{2\AA}
        & \textbf{5\AA}
        & \textbf{8\AA}
        & \textbf{10\AA} & \\
        \midrule
        PiFold
        & 0.2858 & 0.0327 & 0.1763
        & 0.3184 & 0.3752 & 0.1148 \\
        RDesign
        & \textbf{0.3291} & 0.0468 & 0.2227
        & 0.3853 & 0.4566 & 0.0788 \\
        gRNAde
        & 0.3254 & 0.0356 & 0.1938
        & 0.3519 & 0.4298 & 0.0842 \\
        RiboDiffusion
        & 0.2989 & 0.0668 & 0.3742
        & 0.5189 & 0.6102 & \textbf{0.1264} \\
        SOLD
        & 0.2949 & 0.0585 & 0.3913
        & 0.5569 & 0.6321 & 0.1121 \\
        \textbf{GOLD (Dual-Guidance)}
        & 0.3183 & \textbf{0.0724}
        & \textbf{0.4242}
        & \textbf{0.5935}
        & \textbf{0.6670}
        & 0.1015 \\
        \bottomrule
    \end{tabular}
\end{table}

\textbf{Results and Analysis.}
As shown in Table~\ref{tab:designability_diversity}, GOLD achieves the highest structural success rate across all four thresholds, indicating improved structural designability. Meanwhile, GOLD maintains competitive sequence novelty and substantial intra-target diversity, suggesting that its improved structural success does not simply arise from reproducing training sequences. Although its diversity is moderately lower than that of some competing methods, this may reflect a trade-off between structural consistency and sequence diversity under stronger physical guidance. Overall, these results further demonstrate the effectiveness of GOLD for one-to-many RNA inverse folding beyond conventional recovery metrics.

\subsection{Interaction and Sensitivity of Physical Guidance}
\label{app:guidance_sensitivity}

To better understand the behavior of dual-guided sampling, we investigate the interaction between secondary-structure and clash guidance, as well as the sensitivity of the reverse diffusion process to their respective guidance strengths.

\textbf{Interaction Between Guidance Objectives.}
We conduct a two-dimensional parameter sweep over the secondary-structure guidance weight $\omega_{\mathrm{ss}}\in\{0,1,2,3\}$ and clash guidance weight $\omega_{\mathrm{clash}}\in\{0,0.05,0.10,0.50,1.00\}$. Each configuration is evaluated using Sequence Recovery, SS Validity, and Clash Score to examine whether the two guidance objectives cooperate or compete.

\begin{table*}[t]
    \centering
    \caption{Interaction between secondary-structure and clash guidance on the GOLD Benchmark. Each cell reports Sequence Recovery / SS Validity / Clash Score ($\uparrow/\uparrow/\downarrow$). The default configuration is highlighted in bold.}
    \label{tab:guidance_interaction}
    \small
    \setlength{\tabcolsep}{3pt}
    \begin{tabular}{c|ccccc}
        \toprule
        $\omega_{\mathrm{ss}}\backslash\omega_{\mathrm{clash}}$
        & $0$ & $0.05$ & $0.10$ & $0.50$ & $1.00$ \\
        \midrule
        $0$
        & 0.625/0.904/4.51
        & 0.636/0.901/3.11
        & 0.638/0.903/2.76
        & 0.635/0.906/2.28
        & 0.626/0.894/2.77 \\
        $1$
        & 0.627/0.934/4.32
        & 0.637/0.932/3.06
        & \textbf{0.638/0.929/2.67}
        & 0.633/0.928/2.31
        & 0.627/0.927/2.74 \\
        $2$
        & 0.626/0.936/4.43
        & 0.634/0.934/3.08
        & 0.636/0.931/2.69
        & 0.632/0.930/2.27
        & 0.621/0.931/2.81 \\
        $3$
        & 0.627/0.938/4.38
        & 0.635/0.936/3.10
        & 0.637/0.932/2.72
        & 0.630/0.933/2.32
        & 0.624/0.932/2.78 \\
        \bottomrule
    \end{tabular}
\end{table*}

As shown in Table~\ref{tab:guidance_interaction}, the two guidance objectives are largely complementary under moderate weights. Secondary-structure guidance primarily improves topological validity, whereas clash guidance effectively reduces steric conflicts and can also improve sequence recovery. Their joint application largely preserves both benefits without substantially compromising sequence recovery. However, excessively strong clash guidance leads to diminishing structural improvements and eventually degrades sequence recovery, suggesting the importance of balancing the two constraints.

\textbf{Guidance Sensitivity and Reverse-Process Deviation.}
To further understand the effects of guidance strength, we quantify how much each guidance objective perturbs the unguided reverse diffusion process. At reverse step $t\rightarrow s$, we consider the unguided and guided transition distributions:
\begin{equation}
    p_0(z_s\mid z_t)
    =\mathcal{N}(\mu_t^0,\sigma_t^2 I),
    \qquad
    p_g(z_s\mid z_t)
    =\mathcal{N}(\mu_t^g,\sigma_t^2 I).
\end{equation}
Since the two transitions share the same covariance, their step-wise conditional KL divergence is
\begin{equation}
    D_t =
    D_{\mathrm{KL}}
    \left[p_g(z_s\mid z_t)\Vert p_0(z_s\mid z_t)\right]
    =
    \frac{\|\mu_t^g-\mu_t^0\|_2^2}{2\sigma_t^2}.
\end{equation}
Both transitions are evaluated at the same current guided state $z_t$, allowing us to isolate the local perturbation introduced by guidance. We accumulate this divergence along the guided trajectory and normalize it by RNA length $L$ and latent dimension $d$:
\begin{equation}
    KL_{\mathrm{path}}
    =
    \sum_t\frac{D_t}{Ld}.
\end{equation}
The average $KL_{\mathrm{path}}$ over evaluation trajectories provides an operational measure of deviation from the learned reverse process, rather than an exact geometric distance to the latent manifold. We vary one guidance coefficient at a time while setting the other to zero.

\begin{table}[ht]
    \centering
    \caption{Guidance sensitivity and reverse-process deviation under different guidance strengths.}
    \label{tab:guidance_kl_sensitivity}
    \small
    \setlength{\tabcolsep}{3pt}
    \begin{tabular}{llcccc}
        \toprule
        \textbf{Guidance}
        & \textbf{Weight}
        & $\mathbf{KL}_{\mathrm{path}}$
        & \textbf{SS Validity $\uparrow$}
        & \textbf{Clash Score$\downarrow$}
        & \textbf{Seq Recovery $\uparrow$} \\
        \midrule
        SS & $0$ & $0$ & $0.904$ & -- & $0.625$ \\
        SS & $1$ & $0.075$ & $0.934$ & -- & $0.627$ \\
        SS & $2$ & $0.541$ & $0.936$ & -- & $0.626$ \\
        SS & $3$ & $0.905$ & $0.938$ & -- & $0.627$ \\
        \midrule
        Clash & $0$ & $0$ & -- & $4.51$ & $0.625$ \\
        Clash & $0.05$ & $10$ & -- & $3.11$ & $0.636$ \\
        Clash & $0.10$ & $28$ & -- & $2.76$ & $0.638$ \\
        Clash & $0.50$ & $515$ & -- & $2.28$ & $0.635$ \\
        Clash & $1.00$ & $3046$ & -- & $2.77$ & $0.627$ \\
        \bottomrule
    \end{tabular}
\end{table}

As shown in Table~\ref{tab:guidance_kl_sensitivity}, secondary-structure guidance introduces relatively small deviations from the unguided process. Moderate guidance captures most of the improvement in SS validity, while further increasing the weight yields diminishing returns with little effect on sequence recovery. In contrast, clash guidance produces substantially larger trajectory deviations. Moderate clash guidance reduces steric conflicts and improves sequence recovery, whereas excessive guidance sharply increases $KL_{\mathrm{path}}$, accompanied by saturated or deteriorating structural improvements and reduced sequence recovery. These results suggest that overly strong guidance can perturb the learned reverse dynamics beyond a beneficial correction regime. Accordingly, we select the default guidance weights by balancing structural validity, steric feasibility, and sequence recovery.

\section{Analysis of Dual-Guidance Dynamics}
\label{app:guidance_dynamics}

In the main text, global statistical evaluations demonstrated that combining topological and spatial guidance improves the quality of generation. To elucidate the specific physical dynamics underlying the sampling process, this section provides an in-depth case study. We analyze how the guidance parameters $\omega_{ss}$ and $\omega_{clash}$ influence its local spatial frame.

\subsection{Topological Guidance (\texorpdfstring{$\omega_{ss}$}{omega	extunderscore ss})}
\label{app:guidance_ss}

\begin{figure}[ht]
    \centering
    \includegraphics[width=0.9\textwidth]{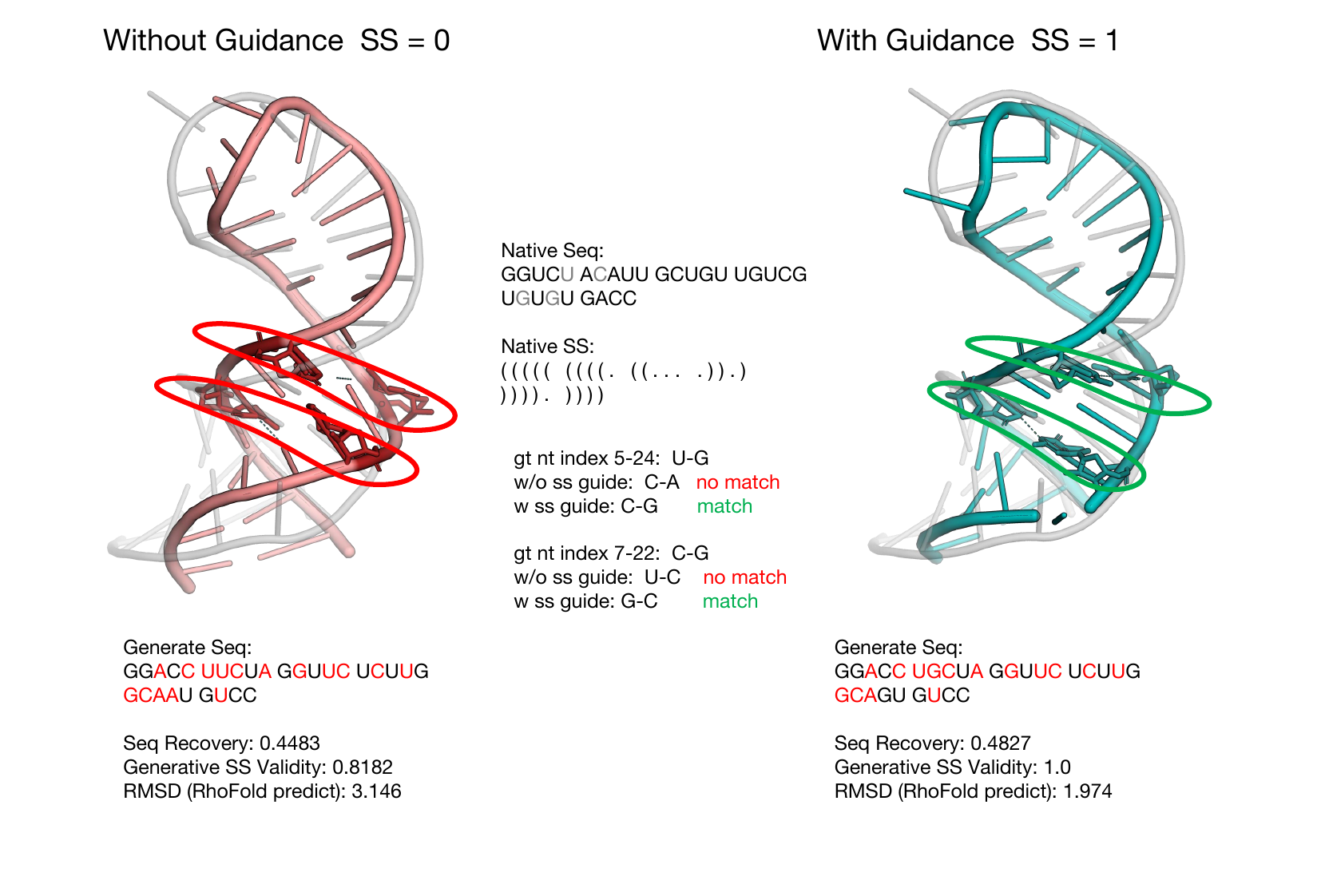} 
    \caption{Observation of topology and sequence intervention under Topological Guidance ($\omega_{ss}$) for target \texttt{2K5Z\_A}. \textbf{Left ($\omega_{ss}=0$):} Without guidance, the generative trajectory leaves the stem partially open, producing unviable base pairs (highlighted in red). \textbf{Right ($\omega_{ss}=1.0$):} The guidance gradient effectively corrects the pairing network by shifting the local sequence probability toward valid Watson-Crick pairs (highlighted in cyan). This intervention successfully closes the stem with 100\% validity, illustrating how the model resolves physical conflicts without simply relying on sequence memorization.}
    \label{fig:guidance_ss}
\end{figure}

Although the main text establishes that $\omega_{ss}$ improves global secondary structure validity, it is important to examine how this guidance alters the generative trajectory at the micro-scale. As illustrated in Figure~\ref{fig:guidance_ss}, in the unguided state ($\omega_{ss}=0$), relying solely on the learned prior of the latent manifold is sometimes insufficient to fully close complex structural motifs, such as the stem of target \texttt{2K5Z\_A}. The absence of explicit spatial pairing constraints leaves the stem partially open, resulting in a pairing validity of 81.82\%.

By introducing the topological guidance weight ($\omega_{ss}=1.0$), the model successfully forms the stem. More importantly, detailed sequence analysis reveals that this structural correction is achieved through context-aware sequence adjustment rather than simply memorizing the native sequence. For example, at the critical pairing site (residues 5 and 24), the unguided model samples an unpaired C-A combination, which physically prevents the stem from forming a stable conformation. Under the gradient guide of $\nabla \mathcal{E}_{ss}$, the sampling trajectory dynamically shifts these specific positions into a stable Watson-Crick pair (C-G). This micro-level correction demonstrates that the guidance gradient effectively reshapes the local sequence probability distribution, actively guiding the generation into valid physical conformations that conforms the target topology.

\subsection{Clash Guidance (\texorpdfstring{$\omega_{clash}$}{omega	extunderscore clash})}
\label{app:clash_ablation}

\begin{figure}[ht]
    \centering
    \includegraphics[width=\textwidth]{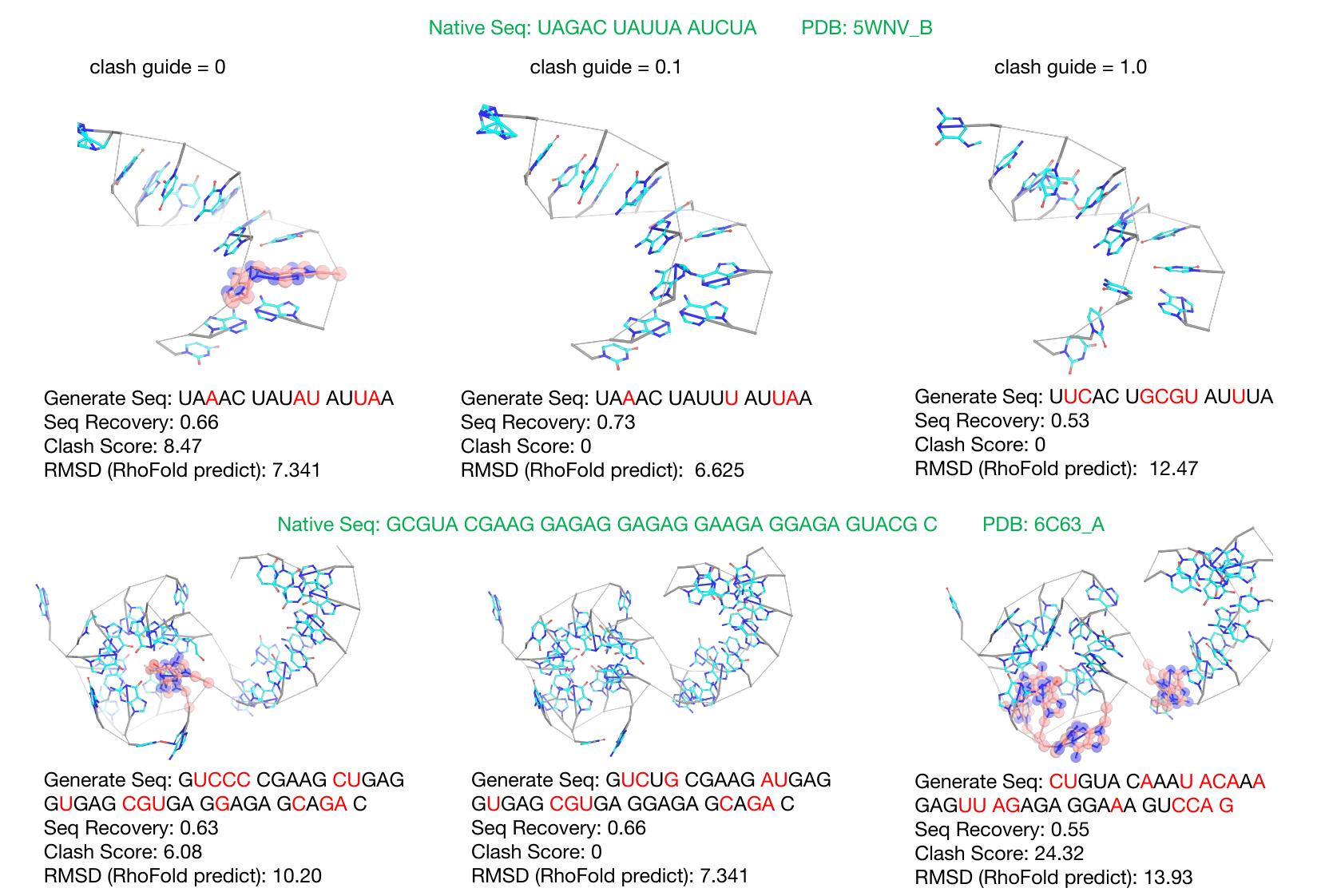} 
    \caption{Ablation of spatial guidance intensity ($\omega_{clash}$) on targets \texttt{5WNV\_B} and \texttt{6C63\_A}. The data demonstrates that while an appropriate weight ($0.1$) resolves physical clashes and preserves sequence recovery, excessive penalization ($1.0$) leads to generative degradation.}
    \label{fig:clash_dynamics}
\end{figure}

\paragraph{Visualization Protocol.} To illustrate the microscopic dynamics of latent spatial guidance, the 3D conformations presented in Figure~\ref{fig:clash_dynamics} and main text Figure ~\ref{fig:case_internal} are generated using the same protocol:
\begin{itemize}[leftmargin=*, topsep=2pt]
    \item \textbf{Raw Generative Output:} The 3D coordinates are directly produced by our diffusion model's sampling trajectory. We intentionally bypassed any external energy minimization or structural relaxation algorithms, ensuring the visualizations reflect the model's unrelaxed generative capabilities under specific guidance weights.
    
    \item \textbf{Topological Backbone Grafting:} Auxiliary backbone atoms absent from our coarse-grained representation were grafted from native structures solely to facilitate continuous PyMOL ribbon rendering, leaving all predicted $C4'$ and side-chain coordinates completely unaltered.
    
    \item \textbf{Standardized Highlighting:} All states are rendered using a fixed camera matrix. Atomic steric overlap, defined as instances where the distance between two non-bonded atoms falls below a threshold of 2.0~\AA, are highlighted using \textbf{red transparent van der Waals spheres} to signify spatial penetration. When clashes are resolved through an appropriate guidance weight, these red spheres disappear, revealing cleanly packed side-chain conformations.
\end{itemize}

\paragraph{Effects of Excessive Penalization.} 
As illustrated in Figure~\ref{fig:clash_dynamics}, scaling the clash guidance reveals a non-linear optimization trade-off. A moderate guidance weight ($\omega_{clash}=0.1$) effectively resolves atomic penetrations while maintaining high sequence recovery. However, applying an excessively high spatial penalty ($\omega_{clash}=1.0$) leads to a decline in generative quality and manifests as two different behaviors:

\begin{itemize}[leftmargin=*, topsep=2pt]
    \item \textbf{Behavior I: Sequence-Structure Trade-off (e.g., Target \texttt{5WNV\_B}).} Under a heavy spatial penalty, the model tends to sacrifice sequence identity to satisfy steric constraints. This altered the sequence prediction, causing the recovery rate to drop from 0.73 to 0.53, and adopts extended side-chain conformations to create spatial gaps. Although this optimization reduces the clash score to zero, the final topology noticeably deviates from the native structure, resulting in an RMSD of 12.47~\AA.
    
    \item \textbf{Behavior II: Gradient-Induced Manifold Deviation (e.g., Target \texttt{6C63\_A}).} In densely packed regions, an excessively repulsive weight produces unusually large gradients during latent sampling. These significant updates can push the latent trajectory $z_t$ outside the learned geometric prior distribution and decoding these out-of-distribution states may disrupt the spatial projection mechanism. Consequently, the collision score increases significantly to 24.32, while the sequence recovery drops to 0.44.
\end{itemize}

These observations suggest that our dual-guided mechanism serves as a physical regularizer. The selected optimal weight ($\omega_{clash}=0.1$) maintains a delicate balance between adhering to the prior geometric manifold and satisfying explicit spatial constraints.

\textbf{Extensibility of Physical Guidance.}
Beyond secondary-structure pairing and steric clashes, GOLD provides a flexible framework for incorporating additional RNA-specific physical interactions. By differentiably reconstructing atomic-level nucleobase coordinates from soft sequence probabilities and glycosidic torsion angles, GOLD allows new physical constraints to be formulated as differentiable energy functions and integrated into the reverse diffusion process without retraining the generative model.

For example, \textbf{base-stacking interactions}, which are important for RNA structural stability, could be modeled using inter-base distances, relative plane orientations, lateral offsets, and geometric overlap. Unlike clash guidance, which primarily penalizes unfavorable atomic overlaps, stacking potentials could encourage favorable spatial arrangements between nucleobases. Similarly, non-canonical base pairing, hydrogen-bond geometry, and base--phosphate interactions could be incorporated through additional energy terms. Although these extensions have not yet been systematically evaluated, they demonstrate the potential of GOLD to accommodate richer physical constraints for RNA inverse folding.

\section{Extended Case Studies}
\label{app:extended_case_study}

To supplement the evaluations in the main text, this section provides detailed structural comparisons. We present two PDB targets, evaluating GOLD against five baselines: PiFold, RDesign, gRNAde, RiboDiffusion, and SOLD.

\subsection{Case Study I}
\label{app:extended_case_study_I}
\begin{figure}[p]
    \centering
    \includegraphics[width=1.0\textwidth]{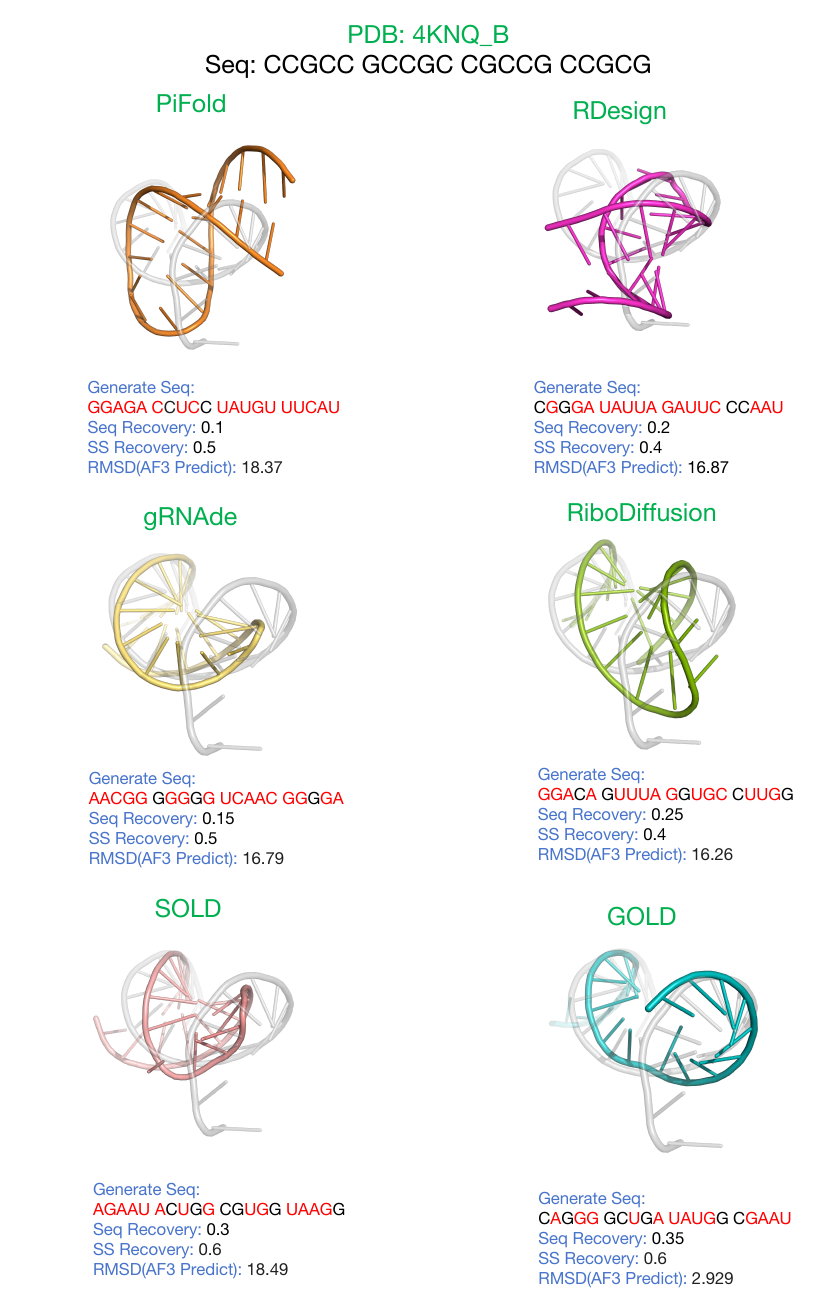}
    \caption{\textbf{Comprehensive Comparison on Target \texttt{4KNQ\_B}.} Baselines exhibit severe topological collapse, whereas GOLD successfully recovers the global fold.}
    \label{fig:case_4knq}
\end{figure}

For the complex target \texttt{4KNQ\_B} (Figure~\ref{fig:case_4knq}), all five baselines fail to orient local fragments globally (RMSD > 16.0~\AA), it highlights the inherent vulnerability of models that remain agnostic to side-chain geometry, which is essential for translating local accuracy into global structural coherence. In contrast, \textbf{GOLD} achieves near-native structural integrity (RMSD \textbf{2.929~\AA}), demonstrating that our joint diffusion process effectively captures the long-range spatial dependencies required for complex 3D assembly.

\subsection{Case Study II}
\label{app:extended_case_study_II}
Target \texttt{6Q8V\_A} is used to evaluate the accuracy of sequence-to-structure prediction. Although baseline methods such as RDesign achieve good sequence recovery, their 3D structures are often mismatched. \textbf{GOLD} overcomes this issue, achieving \textbf{0.94} sequence recovery, \textbf{1.0} SS validity, and a highly accurate 3D conformation.

\begin{figure}[p]
    \centering
    \includegraphics[width=1.0\textwidth]{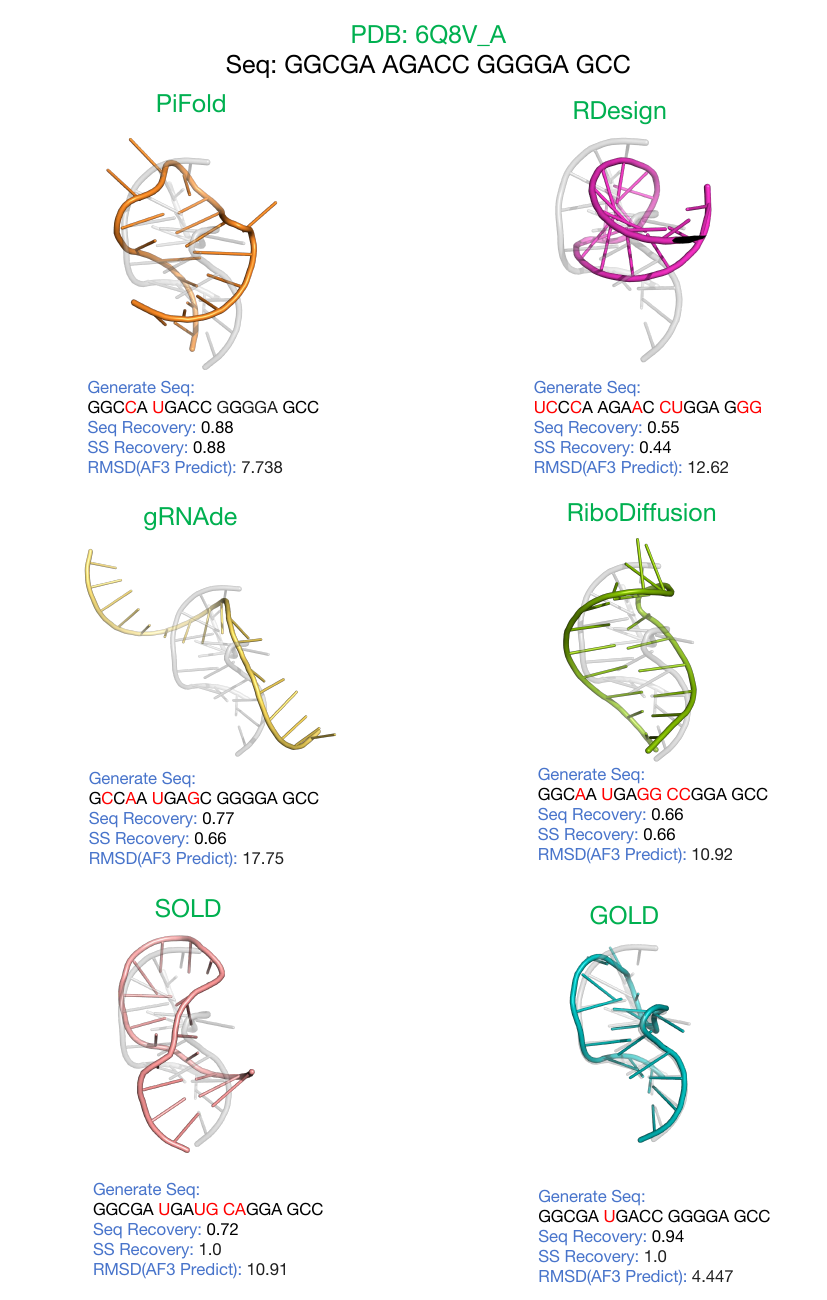}
    \caption{\textbf{Comprehensive Comparison on Target \texttt{6Q8V\_A}.} GOLD demonstrates superior sequence-structure fidelity, achieving 100\% SS validity and the lowest RMSD (4.4\AA) compared to baselines.}
    \label{fig:case_6q8v}
\end{figure}


\newpage
\input{checklist.tex}

\end{document}

%% file: checklist.tex
\section*{NeurIPS Paper Checklist}

\begin{enumerate}

\item {\bf Claims}
    \item[] Question: Do the main claims made in the abstract and introduction accurately reflect the paper's contributions and scope?
    \item[] Answer: \answerYes{} 
    \item[] Justification: In the abstract and introduction, we clearly state the contributions of our paper.

\item {\bf Limitations}
    \item[] Question: Does the paper discuss the limitations of the work performed by the authors?
    \item[] Answer: \answerYes{} 
    \item[] Justification: In main text Conclusion, we clearly describe the limitations of the work.

\item {\bf Theory assumptions and proofs}
    \item[] Question: For each theoretical result, does the paper provide the full set of assumptions and a complete (and correct) proof?
    \item[] Answer: \answerNA{} 
    \item[] Justification: This paper does not include theoretical results.

    \item {\bf Experimental result reproducibility}
    \item[] Question: Does the paper fully disclose all the information needed to reproduce the main experimental results of the paper to the extent that it affects the main claims and/or conclusions of the paper (regardless of whether the code and data are provided or not)?
    \item[] Answer: \answerYes{} 
    \item[] Justification: Appendix C provides the algorithm's details of hyperparameters and workflows. The complete code and
    data are made public as \url{https://github.com/darkflash03/GOLD}.

\item {\bf Open access to data and code}
    \item[] Question: Does the paper provide open access to the data and code, with sufficient instructions to faithfully reproduce the main experimental results, as described in supplemental material?
    \item[] Answer: \answerYes{} 
    \item[] Justification: We clearly discussed the training datasets and other details in Appendix C.

\item {\bf Experimental setting/details}
    \item[] Question: Does the paper specify all the training and test details (e.g., data splits, hyperparameters, how they were chosen, type of optimizer) necessary to understand the results?
    \item[] Answer: \answerYes{} 
    \item[] Justification: The paper specify all the training and test details in Appendix B.

\item {\bf Experiment statistical significance}
    \item[] Question: Does the paper report error bars suitably and correctly defined or other appropriate information about the statistical significance of the experiments?
    \item[] Answer: \answerYes{} 
    \item[] Justification: For both our models and the baselines, we generate multiple outputs for each
    test case and report the average performance across all samples as shown in main text experiment (main results).

\item {\bf Experiments compute resources}
    \item[] Question: For each experiment, does the paper provide sufficient information on the computer resources (type of compute workers, memory, time of execution) needed to reproduce the experiments?
    \item[] Answer: \answerYes{} 
    \item[] Justification: The paper provides sufficient information on the computer resources in Appendix B.3 and Appendix D.2.
    
\item {\bf Code of ethics}
    \item[] Question: Does the research conducted in the paper conform, in every respect, with the NeurIPS Code of Ethics \url{https://neurips.cc/public/EthicsGuidelines}?
    \item[] Answer: \answerYes{} 
    \item[] Justification: The research in this paper conforms in every respect with the NeurIPS Code of Ethics.

\item {\bf Broader impacts}
    \item[] Question: Does the paper discuss both potential positive societal impacts and negative societal impacts of the work performed?
    \item[] Answer: \answerYes{} 
    \item[] Justification:  The paper discussed potential societal impacts in main text conclusion.
    
\item {\bf Safeguards}
    \item[] Question: Does the paper describe safeguards that have been put in place for responsible release of data or models that have a high risk for misuse (e.g., pre-trained language models, image generators, or scraped datasets)?
    \item[] Answer: \answerNA{} 
    \item[] Justification: The paper poses no such risks.

\item {\bf Licenses for existing assets}
    \item[] Question: Are the creators or original owners of assets (e.g., code, data, models), used in the paper, properly credited and are the license and terms of use explicitly mentioned and properly respected?
    \item[] Answer: \answerYes{} 
    \item[] Justification: The existing assets are properly cited and credited.

\item {\bf New assets}
    \item[] Question: Are new assets introduced in the paper well documented and is the documentation provided alongside the assets?
    \item[] Answer: \answerNA{} 
    \item[] Justification: The paper does not release new assets at this point, but will release GOLD Benchmark and code if this paper is accepted.

\item {\bf Crowdsourcing and research with human subjects}
    \item[] Question: For crowdsourcing experiments and research with human subjects, does the paper include the full text of instructions given to participants and screenshots, if applicable, as well as details about compensation (if any)? 
    \item[] Answer: \answerNA{} 
    \item[] Justification: The paper does not involve crowdsourcing nor research with human subjects.

\item {\bf Institutional review board (IRB) approvals or equivalent for research with human subjects}
    \item[] Question: Does the paper describe potential risks incurred by study participants, whether such risks were disclosed to the subjects, and whether Institutional Review Board (IRB) approvals (or an equivalent approval/review based on the requirements of your country or institution) were obtained?
    \item[] Answer: \answerNA{} 
    \item[] Justification: This paper does not involve crowdsourcing nor research with human subjects.

\item {\bf Declaration of LLM usage}
    \item[] Question: Does the paper describe the usage of LLMs if it is an important, original, or non-standard component of the core methods in this research? Note that if the LLM is used only for writing, editing, or formatting purposes and does \emph{not} impact the core methodology, scientific rigor, or originality of the research, declaration is not required.
    \item[] Answer: \answerNA{} 
    \item[] Justification: LLMs were not used as an important, original, or non-standard component of the core methods in this research. They were solely utilized for writing and editing assistance.

\end{enumerate}